\pdfoutput=1

\documentclass[11pt]{article}

\usepackage[]{acl}

\usepackage{times}
\usepackage{latexsym}

\usepackage[T1]{fontenc}

\usepackage[utf8]{inputenc}

\usepackage{microtype}

\usepackage{inconsolata}

\usepackage{xspace}

\usepackage{natbib}
\usepackage{float}
\definecolor{maria-color}{HTML}{7881F2}
\definecolor{andrew-color}{HTML}{3E9CED}
\definecolor{todo-color}{HTML}{ff3232}
\definecolor{joel-color}{HTML}{A52A2A}
\definecolor{light-color}{HTML}{A0A0A0}
\definecolor{story-text-color}{HTML}{783bbb}
\definecolor{story-color}{HTML}{e6d4fa}
\definecolor{non-story-color}{HTML}{cff5f5}

\newcommand{\story}[1]{{\color{story-text-color}[#1]}}
\usepackage{fontawesome5}
\usepackage{graphicx}
\usepackage{fancyhdr}
\usepackage{soul}

\usepackage{scalerel,xparse}
\NewDocumentCommand\telescope{}{
    \scalerel*{
        \includegraphics{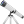}
    }{X}
}

\newcommand{\datasetName}{\telescope StorySeeker\xspace}
\newcommand{\modelName}{\telescope StorySeeker\xspace}
\usepackage{booktabs}
\usepackage{sparklines}

\usepackage{fancyhdr}
\fancypagestyle{firstpage}{
    \fancyhf{} 
    \fancyfoot[C]{
        \thepage\\[0.5em] 
        \hspace*{-\oddsidemargin}\hspace*{-1in}
        \parbox[t]{\paperwidth}{\centering\footnotesize 
            \textit{Proceedings of the 62nd Annual Meeting of the Association for Computational Linguistics (Volume 1: Long Papers)}, pages 7104--7130\\
            August 11-16, 2024 \copyright 2024 Association for Computational Linguistics
        }
    }
    
}

\title{Where Do People Tell Stories Online? \\ Story Detection Across Online Communities}

\newcommand{\aspace}{\hspace{1em}}
\newcommand{\cmu}{$^\diamondsuit$}
\newcommand{\aitwo}{$^\clubsuit$}
\newcommand{\eth}{$^\spadesuit$}
\newcommand{\mcgill}{$^\heartsuit$}

\author{Maria Antoniak\aitwo \aspace Joel Mire\cmu \aspace Maarten Sap\cmu\aitwo \aspace Elliott Ash\eth \aspace Andrew Piper\mcgill 
\vspace{.3em}\\
\small{\aitwo Allen Institute for AI \aspace \cmu Carnegie Mellon University} \aspace \eth ETH Zürich \aspace \mcgill McGill University}

\begin{document}
\maketitle
\begin{abstract}

Story detection in online communities is a challenging task as stories are scattered across communities and interwoven with non-storytelling spans within a single text.
We address this challenge by building and releasing the \mbox{\datasetName} toolkit, including a richly annotated dataset of 502 Reddit posts and comments, a detailed codebook adapted to the social media context, and models to predict storytelling at the document and span levels.
Our dataset is sampled from hundreds of popular English-language Reddit communities ranging across 33 topic categories, and it contains fine-grained expert annotations, including binary story labels, story spans, and event spans. 
We evaluate a range of detection methods using our data, and we identify the distinctive textual features of online storytelling, focusing on storytelling spans. 
We illuminate distributional characteristics of storytelling on a large community-centric social media platform, and we also conduct a case study on \textit{r/ChangeMyView}, where storytelling is used as one of many persuasive strategies, illustrating that our data and models can be used for both inter- and intra-community research.
Finally, we discuss implications of our tools and analyses for narratology and the study of online communities.

\end{abstract}

\section{Introduction}

Automatic detection of stories online is an important but complex task, as online stories can be topically varied and embedded within a longer document.
For example, in Table~\ref{table:annotation-examples}, stories can make up only a small portion of an advice-seeking post, which at first glance may not look like it contains a story.
Successful storytelling detection could enable large-scale analyses of storytelling patterns across online communities, deepening our understanding of the social functions of stories.

Several challenges have hindered the large scale analysis of storytelling online.
First, defining storytelling, i.e., what is a story and what is not a story, is a difficult task that the field of narratology has been concerned with for decades \citep{bal2009narratology}.
Second, unlike traditional media where titles, genres, and other paratextual cues can signal its presence, storytelling in online communities can be fluid and co-occur with many other kinds of discourse \citep[such as seeking information, providing advice;][]{yang2019seekers}. 
Third, existing story datasets are not appropriate for training and evaluating a cross-community story detector; some datasets are not publicly available \citep{ganti-etal-2022-narrative,ganti-findings-2023}, others contain sparse positive labels \citep{gordon2009identifying}, some are not available in English \citep{dos2017portuguese}, some include topical confounders \citep{prize2019hewlett}, and some are largely related to book-based narratives \citep{piper2022toward}.
Finally, no storytelling detection work has tackled detecting stories embedded within documents.

\begin{table}[t]
    \centering
    \small
    \begin{tabular}{@{}p{8cm}@{}}
    \toprule
        The mods \underline{removed} my post last week, very frustrating. Anyway, my major is in Information Science and I'm entering my senior year. \story{I \underline{began} school in CS, but then I \underline{switched} to the iSchool because I \underline{discovered} that the topics were more interesting for me.} I know I shouldn't worry about this, but I feel like my IS degree could hurt my chances of getting into a CS graduate program. I thought you all might have input about my options.
        \\[1ex] 
        \bottomrule
    \end{tabular}
    \caption{
    A motivating example that shows \underline{event} and \story{story} spans and illustrates the difficulty of determining story boundaries and event sequences.
    }
    \label{table:annotation-examples}
\end{table}

\begin{table*}[t]
    \centering
    \small
    \begin{tabular}{@{}p{9.0cm}|p{6.0cm}@{}}
    \toprule
    \textbf{Text} & \textbf{Commentary}\\
    \midrule
    Im trying to lose 4kg over 2 months or 0.5kg a week. The caloric intake calculator says i should aim for about 2560 calories per day. It also breaks it down into three categories carbs (gm) protein (gm) and fats (gm), which it says is 333 147 91 respectively. Am i right in assuming that (gm) is grams because ive only ever seen it (g). And if so is this about accurate? If it is it makes it easier for me because all the foods here say on the label how many grams of these 3 macros are in the food. I understand this might be a bit basic but \story{ive \underline{searched} around the internet and here but \underline{havent} found an answer though ive probs missed it}.
    & Our codebook allows for very short stories composed of as few as two meaningfully connected events. By annotating these minimal spans of narrativity within larger texts, downstream modelling and application can be more attuned to the ways that stories are intermixed with other discourses. Additionally, we can more precisely analyze which textual features are associated with story spans (vs. a less precise approach that considers textual features across the entire text).
    \\[1ex]
    \midrule
    I'm a grad student for context. \story{It \underline{starts} out very general, you \underline{will} spend a year or two really just catching up on literature since, you know, your advisor has been along for the ride for about 15 years so comparatively you're an academic infant. So you \underline{read} a paper which answers questions, but also \underline{creates} more questions for you. But that's ok and is par for the course, so you \underline{look} up more papers which answer those questions, and this is a cycle for your entire career. At some point you \underline{will} think, "why didn't they do this?" but this time... this wondrous time, you \underline{find} there is no further information on the topic. But it's a good thing you \underline{spent} the past two 2 years catching up on it, and now hopefully you \underline{can} try and answer the question yourself. Now you \underline{read} others papers to get answers, which \underline{leads} to more questions. Suddenly it's your own set of questions and answers that \underline{drive} the field.} I am more typical I think and would say after about 12-18 months I \underline{started} asking questions that didn't have really well worked out answers, and that's when the real work begins! I think this diagram is a perfect explanation of how it goes.
    & Although prior work on literary event detection focused solely on realis events \cite{sims-etal-2019-literary}, which are asserted to have actually happened, our codebook recognizes future-tense and hypothetical stories in certain cases, as in this case, where an author's experience informs a story-like sequence of events told in a hypothetical grammatical mode. By accounting for these cases in our codebook, we support a more capacious notion of storytelling, which is not limited to a specific grammatical mode.
    \\[1ex]
    \midrule
    Hey all, guess I'll be coming here more often from now on. I finally \underline{ordered} an Xbox. I also \underline{bought} an HDMI adapter on Amazon. Anything I should know? I \underline{heard} about the undervolting, so I might try that.
    & This example demonstrates that the presence of multiple events does not always imply that there is a story. Stories must contain a \textit{sequence} of events that are meaningfully related.
    \\[1ex]
    \bottomrule
    \end{tabular}
    \caption{Examples of texts that are difficult to classify as either containing a story or not. Our codebook guides our annotations for these edge cases, dealing with issues such as exceptionally short stories, hypothetical stories, and texts with events that are are not clearly interrelated.}
    \label{table:annotation-examples-2}
\end{table*}

To bridge this gap, we formalize the task of story detection by publicly releasing\footnote{\url{https://github.com/maria-antoniak/storyseeker}} \datasetName: the first dataset, codebook, and models for detecting stories and story boundaries across diverse online communities. 
In doing so, (1) we release a public, detailed codebook that can be used across a variety of data types, (2) we explicitly tie our story definition to the related task of \textit{event detection} \citep{sims-etal-2019-literary,vauth2021automated} by annotating event spans, 
(3) our codebook handles token-level boundaries between storytelling and non-storytelling spans, and (4) our data includes large and diverse sets of online communities rather than focusing on a single online setting, furthering our goal of providing a generalizable story annotation framework.

We create the \modelName dataset through iterative rounds of open coding, codebook development, and expert annotation, culminating in  discussions that assign a consensus label to every text in our dataset, drawing on prior work in narratology \citep{herman2009basic,piper-etal-2021-narrative}.
The final dataset includes expert-annotated Reddit posts and comments with story labels, story boundaries, and event labels.
This dataset ranges across hundreds of popular subreddits drawn from 33 broad topic categories, resulting in 235 storytelling documents and 1,739 event-spans over 502 English-language Reddit posts and comments.

\datasetName opens up new research questions for computational story analysis.
While prior work has focused on storytelling in specific communities and topics \citep[e.g., healthcare communities;][]{ganti-etal-2022-narrative,ganti-findings-2023}, many larger research questions about online storytelling require the ability to detect stories across large and diverse sets of communities.
Prior work has not been able to answer important questions such as where storytelling happens online, which community features lead to more storytelling, or how storytelling is used rhetorically across social contexts ---
and until now, resources have not existed to support such research.

Our contributions include the following.
\begin{itemize}
    \item We release the \modelName dataset, codebook, and models to detect storytelling documents and text spans.
    \item Using \modelName, we quantify storytelling rates across many online communities for the first time, finding that the prevalence of storytelling varies widely, ranging from low storytelling rates in religion-focused communities to high storytelling rates in healthcare-focused communities. 
    \item We identify text features that are more frequently present in storytelling spans.
    \item We map communities by their storytelling rates and the distinctiveness of their stories, providing insights for researchers interested in studying particular community categories. 
    \item We illustrate the effectiveness of our toolkit not only for inter-community analyses but also demonstrate how storytelling can be examined as a rhetorical strategy in a specific community through a case study of \textit{r/ChangeMyView}.
\end{itemize}

\section{Related Work}
\label{section:related-work}

\paragraph{Narratology}
Storytelling is a broad concept that has been explored by fields as diverse as economics \citep{shiller2020narrative}, literary theory \citep{bal2009narratology}, sociology \citep{berger2004storytelling}, and NLP \citep{eisenberg-finlayson-2017-simpler, piper-etal-2021-narrative, ranade2022computational}. 
Arriving at agreement on a single story definition is a challenging task.

In the field of narrative theory (``narratology''), storytelling has been defined by its emphasis on sequences of events, aspects of change and/or conflict, and embodiment or ``feltness'' \cite{herman2009basic, bruner1991narrative, fludernik2002towards}. As \citet{herman2009basic} writes, ``Narrative roots itself in the lived, felt experience of human or human-like agents interacting in an ongoing way with their surrounding environment.'' \citet{piper-etal-2021-narrative} propose a minimum schema for capturing storytelling that emphasizes the presence of ten basic elements, including an agent, action, and location in time and space.

\paragraph{Storytelling datasets}
Operationalizing storytelling schemas for NLP is an active area of research \citep{roos2021narratives, piper2021detecting, piper2022toward}.
NLP research on narratives and storytelling has deployed a wide range of definitional dimensions, the most common being sequences of events arranged temporally, causally related events leading to resolutions, and the presence of entities or characters, while a smaller number include a rhetorical purpose for the text and world building \citep{ceran2012semantic,yao-huang-2018-temporal,eisenberg-finlayson-2017-simpler,castricato-etal-2021-fabula,alzahrani2016story}.
See \ref{appendix-subsection-prior-definitions} for an enumeration of story definition features used in prior work.

Prior work in story annotation has mostly focused on specific discourse domains such as healthcare \citep{ganti-etal-2022-narrative}, argumentation \citep{falk-lapesa-2022-reports,falk-lapesa-2023-storyarg}, book publishing \citep{piper2022toward}, bereavement \citep{doyle2024stories}, or blogs \citep{dos2017portuguese,gordon2009identifying}.
To our knowledge, all prior work on story detection has focused on passage- or document-level annotations, except for a set of preliminary annotation guidelines for embedded narratives without an associated dataset by \citet{Eisenberg2021Narrative} and a domain-specific study by \citet{falk-lapesa-2023-storyarg}.
Much past annotation work is not publicly available \citep{ceran2012semantic,ganti-findings-2023}, due to data agreements, sensitive content, and other constraints.

We expand on these works by widening our view to many communities and topics, by using a span-based approach, and by publicly releasing our annotations.
Our annotation guidelines presented in \S\ref{section:codebook} most closely resemble the guidelines provided by \citet{eisenberg-finlayson-2017-simpler} and \citet{Eisenberg2021Narrative}, which rely on events and characters, and draw most inspiration from the narrative annotation guidelines provided by \citet{piper2021detecting} and the event annotation guidelines provided by \citet{sims-etal-2019-literary}.

\paragraph{Story detection}
Most prior work on automatic story detection has focused on using feature-based classification approaches, relying on features like n-grams, POS tags, and coreference chain length
\citep{ceran2012semantic,gordon2009identifying,yao-huang-2018-temporal,eisenberg-finlayson-2017-simpler, piper2021detecting}.
See \ref{appendix-subsection-prior-definitions} for an enumeration of story features used for prediction in prior work.
These studies either had an explicit goal of using interpretable methods or were conducted prior to the arrival of large language models (LLMs).
Two newer works have attempted narrative detection using LLMs \citep{ganti-etal-2022-narrative,ganti-findings-2023}. 
Both studies hand-annotate a small set of texts from online healthcare communities with binary story labels, and both find that fine-tuned BERT-based models perform better than classical models at the document-level story prediction task.

\section{Designing a Story Codebook}
\label{section:codebook}

Our guidelines must work across diverse online communities while recognizing features that make storytelling in these contexts potentially unique from storytelling in literary or visual contexts.

Our interdisciplinary team worked iteratively to code data, compare labels, and design a codebook containing a story description that fit our context (diverse online communities) and research goal (measuring storytelling).
Drawing from prior work, we focused our story identification guidelines on agent-centered events, leading to a two-step annotation process: first, the annotation of event spans, and second, the annotation of story spans.
We provide our full codebook instructions in \ref{appendix-section-full-codebook}.

Our strong emphasis on events while annotating story spans is novel and adds consistency to the annotation of a diverse set of texts.
It also allows us to build on prior work on event span annotation \citep{sims-etal-2019-literary} by reconsidering this task within the context of storytelling, leading us to a more flexible description of events than prior work.

Importantly, our codebook should be considered not as presenting complete or final definitions of either stories or events but rather providing actionable advice that captures features useful for the consistent annotation of these phenomena.

\paragraph{Story annotation guidelines} 
During our story annotation process, we look for texts containing \textit{a sequence of events involving one or more people.} 
Notably, as long as a text meets these requirements, we annotate it as a story, regardless of whether it is as short as one sentence or as long as the entire post.
Storytelling is thus not limited to only posts that are entirely focused on storytelling.
Unlike some prior work, we do not include world-building or setting as features in our guidelines, as such descriptions are rare in our data, and we also do not include the presence of a narrator, as all Reddit posts and comments by default have their authors as narrators.
When selecting story spans, we do not include the post title, introductory text about the subreddit, or explanations and discussions external to the story, but we do include non-event text that sets the stage, summarizes the story, or ends with a lesson learned.

\paragraph{Event annotation guidelines} 
Our events guidelines draw heavily from the guidelines in \citet{sims-etal-2019-literary}, summarized as \textit{an event is a singular occurrence at a particular place and time}.
We modify this guideline in three ways: (1) We do not require verbs to be in the past tense, (2) we sometimes allow hypothetical verbs to be labeled as events, and (3) we sometimes allow verbs with negation to be labeled as events.
These changes reflect both the different norms in online communities (e.g., the frequency of using present tense to tell stories), and the story-detection intention underlying our event labeling (e.g., the importance of some negative events in a story sequence). 
Some changes, like the increased attention to stative verbs, reflect choices also made by \citet{vauth2021automated}, a follow-up work on literary event detection.

\section{Creating a Multi-Community Corpus}
\label{section:corpus}

We developed the \datasetName dataset, a collection of Reddit posts that introduces the story span annotation task and covers a large number of online communities.
To detect storytelling across diverse settings, we required training and evaluation data that contained both storytelling and non-storytelling communication mixed together by topic, context, and even within a single document.

\paragraph{Data source}
\label{subsection:data-source}

We sample texts from Webis-TLDR-17, a corpus of 3.8 million Reddit posts and comments \citep{volske-etal-2017-tl}. 
This dataset was designed for summarization research, and so unlike a random sampling of Reddit data, each text in this dataset contains contentful texts (texts with coherent sentences leading to a summary statement, rather than images, links, or other kinds of texts). 

Sampling across the 500 most frequent subreddits in the dataset, we follow an open-coding approach to categorize the subreddits into 33 categories (e.g. \textit{professional advice}).
We use these categories to remove sensitive (e.g., \textit{r/confessions}), toxic (e.g., \textit{r/pettyrevenge}), explicit (e.g., \textit{r/sex}), and non-English (e.g., \textit{r/mexico}) subreddits from the human annotation tasks, and we use these categories to structure our analyses (\S\ref{section:analysis}).
We show the full set of categories and their member subreddits in \ref{appendix-section-additional-info-dataset}.
We sample a balanced set of five texts from each of the filtered subreddits, downsampling \textit{gaming}\footnote{The \textit{gaming} category has twice the number of subreddits of the next most frequent category, \textit{hobbies} (100 versus 49).} and requiring that each text contains at least 100 tokens and no more than 500 tokens.

\paragraph{Annotation process}

Our annotation process included highlighting both story and event spans, as we found that first identifying events was crucial in making the story span decision. 
After several rounds of codebook construction, two of the authors independently annotated the target data using Prodigy; each annotator annotated the full set of texts.\footnote{\url{https://prodi.gy/}} 
Annotation was challenging, as texts can contain many events, and we included frequent rounds of group deliberation for difficult examples as well as a final round of discussion to arrive at a single consensus label for each document.

\begin{table}[t]
    \centering
    \scriptsize
    \begin{tabular}{@{}p{0.8cm}p{1.3cm}p{0.8cm}p{0.8cm}p{2cm}@{}}
    \toprule
    \multicolumn{5}{c}{\textit{\textbf{Story} Annotation: $N$=502, Cohen's $k$=0.66 (binary), 0.72 (span)}} \\
    \midrule
    \textbf{Type} & \textbf{Story Label} & \textbf{\# Docs} & \textbf{\% Docs} & \textbf{\# Tokens / Span}  \\
    \midrule
    post
    &
    \colorbox{story-color}{story}
    & 137
    & 64\%
    & 119 (mean)
    \\[1ex]
    &
    \colorbox{non-story-color}{non-story}
    & 78
    & 36\%
    & --
    \\[1ex]
    comment
    &
    \colorbox{story-color}{story}
    & 98
    & 34\%
    & 92 (mean)
    \\[1ex]
    &
    \colorbox{non-story-color}{non-story}
    & 189
    & 66\%
    & --
    \\[2ex]
    \bottomrule
    \end{tabular}
    \caption{Overview of the annotated data.}
    \label{table:annotations-overview}
\end{table}

\paragraph{Final corpus}

Our final \datasetName dataset includes 502 texts with event- and story-spans (see Table \ref{table:annotations-overview}), sampled randomly from the larger dataset and removing eight toxic texts that escaped our automatic filters.
Our inter-annotator agreement for both spans is in the traditional ``substantial'' range (Cohen's $k$ = 0.65 for event spans, 0.72 for story spans).
47\% of the texts included story spans according to the final consensus labels.

\section{Developing a Story Detection Model}
\label{section:storydetection}

\begin{table*}[h]
    \centering
    \scriptsize
    \begin{tabular}{@{}p{2.9cm}|p{1.3cm}@{\hspace{.75em}}p{1.3cm}@{\hspace{.75em}}p{1.3cm}|p{1.3cm}@{\hspace{.75em}}p{1.3cm}@{}@{\hspace{.75em}}p{1.3cm}@{}}
    \toprule
    \textbf{Model} & \textbf{P} & \textbf{R} & \textbf{F1} & \textbf{P} & \textbf{R} & \textbf{F1} \\
    \midrule
    \textit{Document Classification} & \multicolumn{3}{c}{\textit{Story}}  |& \multicolumn{3}{c}{\textit{\textbf{Not} Story}} \\
    \midrule
    \textbf{SVM with TF-IDF}    
    & $0.82 \color{light-color}\pm 0.06$
    & $0.69 \color{light-color}\pm 0.11$
    & $0.74 \color{light-color}\pm 0.08$
    & $0.77 \color{light-color}\pm 0.06$
    & $0.87 \color{light-color}\pm 0.03$
    & $0.81 \color{light-color}\pm 0.03$
    \\[1ex]
    \textbf{Fine-tuned RoBERTa} 
    & $0.87 \color{light-color}\pm 0.02$
    & $0.85 \color{light-color}\pm 0.08$
    & $0.86 \color{light-color}\pm 0.04$
    & $0.88 \color{light-color}\pm 0.05$
    & $0.89 \color{light-color}\pm 0.03$
    & $0.88 \color{light-color}\pm 0.02$
    \\[1ex]
    \textbf{GPT-4 Zero-Shot} 
    & $0.83 \color{light-color}\pm 0.02$
    & $0.76 \color{light-color}\pm 0.06$
    & $0.79 \color{light-color}\pm 0.03$
    & $0.80 \color{light-color}\pm 0.04$
    & $0.87 \color{light-color}\pm 0.02$
    & $0.83 \color{light-color}\pm 0.02$
    \\[1ex]
    \textbf{GPT-4 Few-Shot} 
    & $0.84 \color{light-color}\pm 0.05$
    & $0.70 \color{light-color}\pm 0.06$
    & $0.76 \color{light-color}\pm 0.03$
    & $0.77 \color{light-color}\pm 0.04$
    & $0.88 \color{light-color}\pm 0.04$
    & $0.82 \color{light-color}\pm 0.03$
    \\[1ex]
    \textbf{GPT-4 C-o-T} 
    & $0.56 \color{light-color}\pm 0.05$
    & $0.96 \color{light-color}\pm 0.03$
    & $0.71 \color{light-color}\pm 0.04$
    & $0.91 \color{light-color}\pm 0.06$
    & $0.35 \color{light-color}\pm 0.02$
    & $0.50 \color{light-color}\pm 0.03$
    \\[1ex]
    \textbf{GPT-3.5-Turbo Zero-Shot} 
    & $0.93 \color{light-color}\pm 0.02$
    & $0.37 \color{light-color}\pm 0.08$
    & $0.53 \color{light-color}\pm 0.08$
    & $0.64 \color{light-color}\pm 0.04$
    & $0.97 \color{light-color}\pm 0.01$
    & $0.77 \color{light-color}\pm 0.03$
    \\[1ex]
    \textbf{GPT-3.5-Turbo Few-Shot} 
    & $0.79 \color{light-color}\pm 0.06$
    & $0.53 \color{light-color}\pm 0.10$
    & $0.63 \color{light-color}\pm 0.09$
    & $0.68 \color{light-color}\pm 0.07$
    & $0.88 \color{light-color}\pm 0.03$
    & $0.77 \color{light-color}\pm 0.05$
    \\[1ex]
    \textbf{GPT-3.5-Turbo C-o-T} 
    & $0.56 \color{light-color}\pm 0.07$
    & $0.92 \color{light-color}\pm 0.03$
    & $0.69 \color{light-color}\pm 0.05$
    & $0.85 \color{light-color}\pm 0.06$
    & $0.36 \color{light-color}\pm 0.07$
    & $0.50 \color{light-color}\pm 0.05$
    \\[1ex]
    \midrule
    \textit{Token Classification} & \multicolumn{3}{c}{\textit{Story ($N=7,756$)}}  |& \multicolumn{3}{c}{\textit{\textbf{Not} Story ($N=20,477$)}} \\
    \midrule
    \textbf{Fine-tuned RoBERTa} 
    & $0.77 \color{light-color}\pm 0.05$
    & $0.79 \color{light-color}\pm 0.08$
    & $0.78 \color{light-color}\pm 0.04$
    & $0.90 \color{light-color}\pm 0.03$
    & $0.88 \color{light-color}\pm 0.03$
    & $0.89 \color{light-color}\pm 0.01$
    \\[0.5ex]
    \textbf{GPT-4 Few-Shot} 
    & $0.52 \color{light-color}\pm 0.06$
    & $0.86 \color{light-color}\pm 0.04$
    & $0.64 \color{light-color}\pm 0.05$
    & $0.88 \color{light-color}\pm 0.04$
    & $0.57 \color{light-color}\pm 0.04$
    & $0.69 \color{light-color}\pm 0.04$
    \\[0.5ex]
    \bottomrule
    \end{tabular}
    \caption{Cross-validation results ($k=5$) for document and token classification performance across methods, broken apart by the binary labels where the presence of a story is determined using the \textit{consensus} label. 
    For each score, we show the mean and standard deviation across the $k$ folds.
    We used GPT model versions gpt-4-0314 and gpt-3.5-turbo-0613; see \ref{appendix-section-gpt-prompts} for GPT prompts. 
    } 
    \label{table:classification-performance}
\end{table*}

\begin{figure}[t]
    \centering
        \centering
        \includegraphics[width=0.6\linewidth]{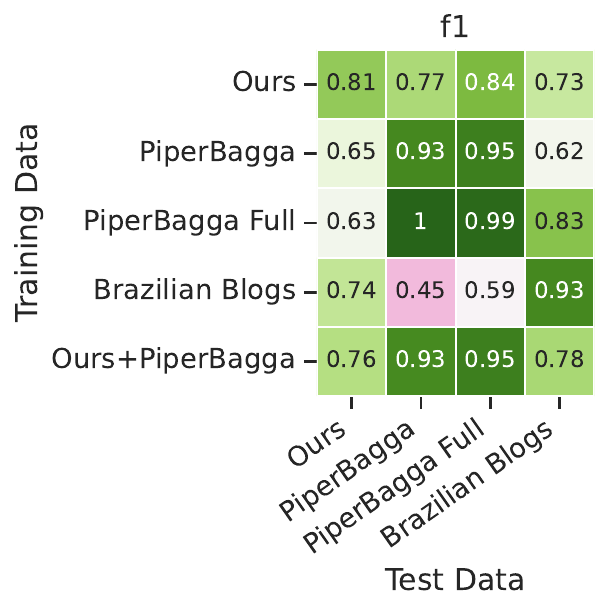}
    \caption{Comparison of document classification performance across datasets using RoBERTa model fine-tuned on our \textit{consensus} labels or the narrative labels in the associated datasets, which were split into 70-30 finetuning and test sets. Cells show F1 scores for the story label.}
    \label{figure:heatmap-generalization}
\end{figure}

We release two fine-tuned classifiers as part of \modelName to detect storytelling documents and spans across diverse online communities.

\paragraph{Prediction methods}

To predict the presence of storytelling, we evaluate a variety of prediction models and methods to cover reasonable baselines that vary in cost, GPU usage, and need for labeled data.
These methods include a SVMs baseline using TF-IDF features, 
a fine-tuned RoBERTa\footnote{We used the Hugging Face library with the \texttt{roberta-base} model for both document and sequence prediction, using \texttt{RobertaForSequenceClassification} for document prediction and \texttt{RobertaForTokenClassification} for sequence prediction from the \texttt{transformers} library.} model \citep{liu2019roberta}, and zero-shot and few-shot GPT-4 prompting \citep{openai2023gpt4}. 

For the document classification task, we predict the binary presence of storytelling in the document, while for the span detection task, we predict whether each token is part of a story.
We divide our expert-annotated data into a training/prompting set of 301 texts, a validation set of 100 texts, and a test set of 101 texts.
We rely on the consensus labels, i.e., if the annotators agree after discussion that a text contains a story span, then we use this as either (i) a positive instance of storytelling or (ii) an indicator of whether to include the union of the annotator's spans for token prediction.

We experimented with different prompts using our validation set, including variations of the below task that included examples, chain-of-thought questions, and guidelines.
\begin{quote}
    \setlength{\parindent}{0cm}\ttfamily{
    Task: A story describes a sequence of events involving one or more people. Does the following text contain a story? Answer yes or no, and then explain
    your reasoning.\newline
    Text: <TEXT>\newline
    Answer:
    }
\end{quote}
For more details and the full set of prompts, see \ref{appendix-section-gpt-prompts}.
For few-shot tests with OpenAI models, we interleave two positive and two negative examples in the prompt, using model versions GPT-4-0314 and GPT-3.5-turbo-0613.

\paragraph{Evaluation results}
We find the best overall story-detection performance from the finetuned RoBERTa model (Table \ref{table:classification-performance}), but the GPT-4 prompts are sometimes comparable.
GPT-4 performed better than GPT-3.5 and chain-of-thought prompting \citep{camburu2018snli} did not yield consistent improvements.
Given our small expert set, these results are averaged over $k=5$ cross-validation folds, and we show both the means and standard deviations.
We examine model errors in \ref{appendix-subsection-error-analysis}.

\paragraph{Cross-dataset prediction performance}

Overall, \datasetName performs better on our Reddit data than identical classifiers fine-tuned using other story datasets, demonstrating the importance of creating a customized codebook, dataset, and model.
Figure \ref{figure:heatmap-generalization} shows the results across the story-annotated datasets to which we could reasonably compare.
We find that our training data generalizes to these non-Reddit datasets reasonably well, with the lowest performance for the set of Brazilian blog posts \citep{dos2017portuguese}, a dataset that we automatically translated from Portuguese to English using Google Translate for this experiment.
The PiperBagga datasets \citep{piper2022toward} include  both a full dataset, with a large number of automatic genre-based labels, and a smaller dataset containing 394 hand-annotated texts mostly from literary sources.
Storytelling in the PiperBaggas datasets is strongly correlated with text genre, resulting in models that are overfit when finetuned and tested on the same datasets.

We do not include comparisons to the ICWSM 2009 Spinn3r Dataset \citep{The_ICWSM_2009_Spinn3r_Dataset,gordon2009identifying} because of its very low ratio of stories to non-stories, nor do we include comparisons to the Hewlett Essays \citep{prize2019hewlett} because of their strong topical confounders (storytelling is almost perfectly correlated with essay prompt).
Other datasets were not accessible for comparison.
None of the other datasets include span annotations, and so we cannot draw comparisons about story spans.

\section{Analysis}
\label{section:analysis}

\begin{figure}[t]
    \centering
        \centering
        \includegraphics[width=\linewidth]{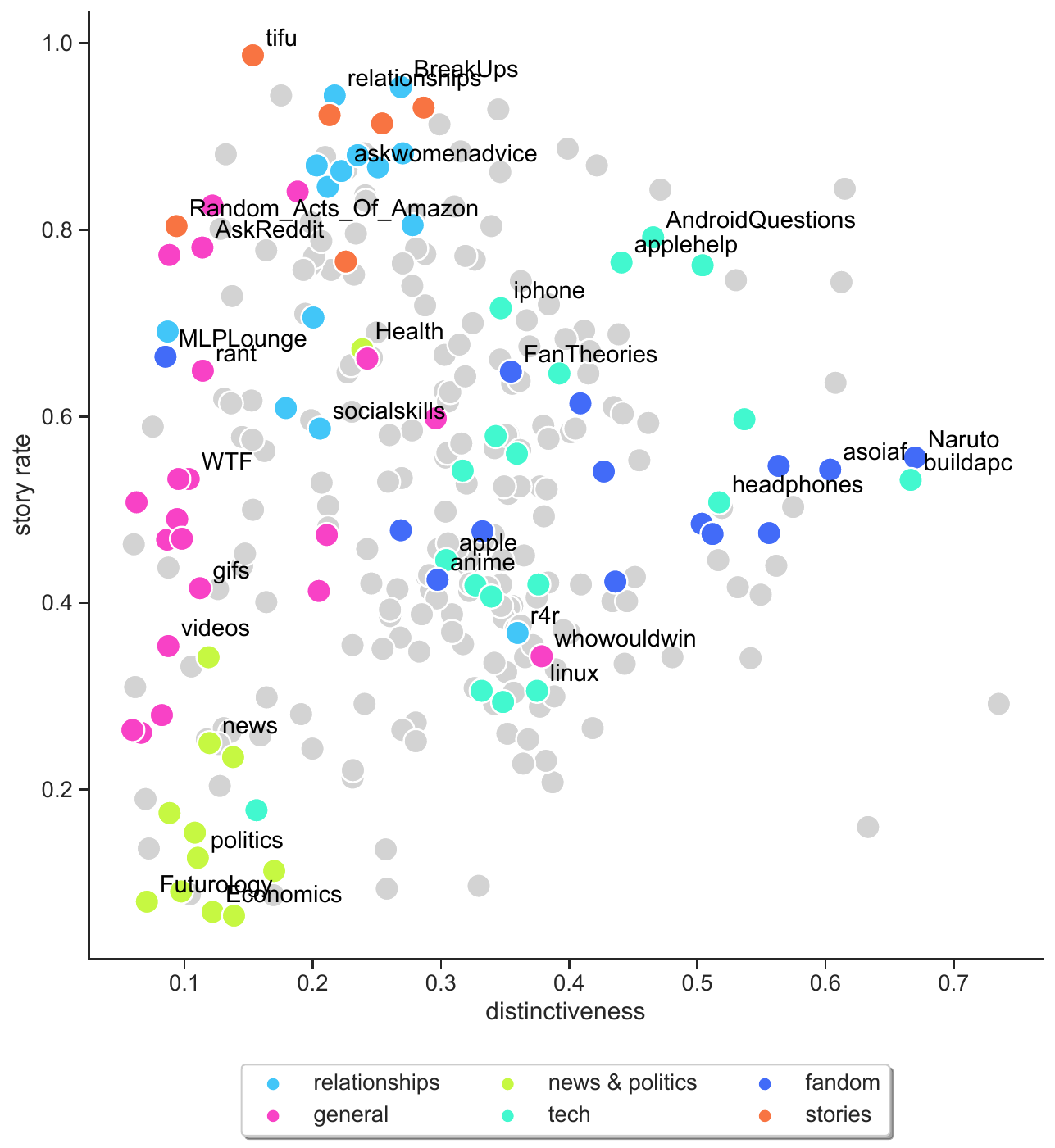}
    \caption{Some subreddits contain stories on specific topics, while others draw a wide range of storytelling topics. We show the 500 most frequent subreddits, colored by category. The y-axis shows the rate of storytelling in the subreddit (predicted by our classifier), and the x-axis shows the distinctiveness of the subreddit's vocabulary (calculated only for texts containing stories). See Figure \ref{figure:boxplot-categories} in the Appendix to see overall category rankings and variation.}
    \label{figure:scatterplot-categories}
\end{figure}

\subsection{What are the signs of storytelling spans?}
\label{section:what-are-signs}

\begin{table}[t]
    \centering
    \scriptsize
    \begin{tabular}    {@{}p{2.9cm}|p{1.0cm}p{1.0cm}p{1.2cm}}
    \toprule
    \textbf{Measure} & \textbf{Effect Size ($d$)} & \textbf{Direction} & \textbf{$p$-value}\\
    \midrule
    expert-annotated events
    & 2.122***
    & \colorbox{story-color}{story}
    & $p<0.001$
    \\[1ex]
    past tense
    & 1.742***
    & \colorbox{story-color}{story}
    & $p<0.001$
    \\[1ex]
    \textit{realis} events
    & 1.655***
    & \colorbox{story-color}{story}
    & $p<0.001$
    \\[1ex]
    1st-person singular pronouns
    & 1.051***
    & \colorbox{story-color}{story}
    & $p<0.001$
    \\[1ex]
    3rd-person singular pronouns
    & 0.459***
    & \colorbox{story-color}{story}
    & $p<0.001$
    \\[1ex]
    entity mentions
    & 0.346***
    & \colorbox{story-color}{story}
    & $p<0.001$
    \\[1ex]
    concreteness
    & 0.329***
    & \colorbox{story-color}{story}
    & 0.001
    \\[1ex]
    non-past tense
    & 1.296***
    & \colorbox{non-story-color}{non-story}
    & $p<0.001$
    \\[1ex]
    is comment (vs. post)
    & 0.612***
    & \colorbox{non-story-color}{non-story}
    & $p<0.001$
    \\[1ex]
    2nd-person pronouns
    & 0.551***
    & \colorbox{non-story-color}{non-story}
    & $p<0.001$
    \\[1ex]
    sentence length
    & -
    & -
    & 0.345
    \\[1ex]
    \textit{1st-person plural pronouns}
    & --
    & --
    & 0.345
    \\[1ex]
    \bottomrule
    \end{tabular}
    \caption{Results of $t$-tests comparing features between texts labeled as containing stories vs. not containing stories in the \datasetName dataset. The story group is composed solely by the story spans, as opposed to the entire text labeled as containing a story. We control for multiple comparisons using the Holm method (***: $p<0.001$; **: $p<0.01$; *: $p<0.05$).}
    \label{table:story-features}
\end{table}

Based on prior literature and our iterations of annotation and codebook refinement, we identify a set of features that we expect may be associated with storytelling in social media. These include entity and pronoun rates \citep{eisenberg-finlayson-2017-simpler, piper2022toward}, events \cite{huhn2009event, gius2022towards, sap-pnas-2022, sims-etal-2019-literary}, verb tense, concreteness \citep{piper2022toward, Brysbaert_Warriner_Kuperman_2013}, whether a text is a post or comment, and average sentence length. 
See \ref{appendix-section-story-feature-analysis} for context on prior work related to these features, the precise definitions we adopted for this study, and a feature comparison test with another narrative detection dataset composed of mostly literary texts.

Using our consensus labels to compare story spans with non-storytelling texts, we test whether each feature is more prominent in one group than the other. Specifically, we run t-tests on the features, applying the Holm method \citep{holm1979simple} to account for multiple comparisons (see Table \ref{table:story-features}). 
The tests indicate that, in decreasing order of effect size, the frequency of our expert-annotated events, past-tense verbs, \textit{realis} events \cite{sims-etal-2019-literary}, first-person singular pronouns, third-person singular pronouns, entity mentions, and concrete terms are significantly more frequent in stories. 
Conversely, present- and future-tense verb tenses, the text type being a comment (instead of a post), and second-person pronouns are significantly less frequent in stories.

\subsection{Where do people tell stories?}
\label{section:where-stories}

We use the fine-tuned RoBERTa model, which we release as part of \datasetName, to identify storytelling across a larger set of texts from the Webis-TLDR-17 dataset, sampling a balanced set of 1k texts at random from each subreddit that contains at least that number of texts, resulting in a set or 291 subreddits for prediction.
For this experiment, we assign each text a binary prediction for the presence of storytelling.

Using our predicted story labels, we find meaningful differences in the rate at which individuals tell stories across different communities.
We find a high of 0.98 stories per all posts and comments (\textit{r/tifu}, i.e., \textit{Today I F*-ed Up}) and a low of 0.11 (\textit{r/Futurology}). 
Overall, 52\% of all the texts were predicted to contain stories, with a macro average across categories of 58\%, suggesting that storytelling is indeed a frequent communicative behavior across different kinds of communities.

Figure \ref{figure:boxplot-categories} shows the subreddit categories ranked by their storytelling rates. 
The \textit{stories}, \textit{addiction}, \textit{animals}, and \textit{healthcare} categories are ranked highest, while \textit{countries}, \textit{news \& politics}, \textit{software dev}, and \textit{religion} are ranked lowest; these categories include subreddits whose mean storytelling rate is relatively low.
Some categories, e.g., \textit{professional advice}, have wide variation in the storytelling rates of their subreddits. In general, storytelling is more prevalent within communities focused on personal issues related to health and relationships.

When interpreting these results, it is important to keep in mind that our comparisons are only across texts in the Webis-TLDR-17 dataset, i.e. \textit{coherent} texts with summaries. 
Since many subreddits predominately include photos or other content, our storytelling rates should be interpreted as relative rankings rather than as absolute rates of storytelling.
We provide additional results in Figure \ref{figure:barplot-sd-subreddits} in \ref{appendix-section-additional-results} that are not restricted to texts with summaries, validating that general patterns hold when sampling over the full space of subreddit posts.

\subsection{How distinctive are stories by community?}\label{section:how-distinctive}

Drawing on work on dataset cartography \citep{swayamdipta-etal-2020-dataset}, we calculate the \textit{distinctiveness} of the stories shared in different communities to help researchers make decisions about story detection methods for specific subsets of communities.
Distinctiveness measures how similar story vocabulary is in comparison to a background vocabulary distribution and has been used in prior work to map Reddit and scholarly communities \citep{zhang2017community,lucy-etal-2023-words}.
Following \citet{zhang2017community}, we first calculate the specificity $S$ of each word used in a community,
\begin{equation}
    S_c(w) = log \frac{P_c(w)}{P_C(w)}
\end{equation}
where the score compares the probability of each word ($w$) in a single subreddit ($c$) versus its probability across all of the subreddits ($C$).
To measure differences in storytelling behavior, we average the specificity scores across the vocabulary, arriving at a single \textit{distinctiveness} score for each subreddit.
Importantly, we calculate distinctiveness only for the texts predicted as containing stories because we are interested in the language used in stories, not in the subreddit overall.

In Figure \ref{figure:scatterplot-categories}, we map communities across two axes: their \textit{story rate}, predicted by our story detector, and the \textit{distinctiveness} of their vocabulary.
Categories of subreddits form interpretable clusters.
We show 291 subreddits (those matching our filtering criteria, see \S\ref{section:storydetection}), and Table \ref{table:matrix-specificity-storytelling} in  \ref{appendix-section-additional-results} shows the ``corners'' of this plot, i.e., the subreddits with the most and least storytelling and distinctiveness.
For example, subreddits in the \textit{stories} category, such as \textit{r/Glitch\_in\_the\_Matrix}, tend to have both high rates of storytelling and low distinctiveness --- these subreddits elicit stories that do not use consistently distinctive language.
Some categories contain a wide range, e.g., the \textit{technology} category contains subreddits varying across half the distinctiveness range, from \textit{r/apple} (not at all distinctive) to \textit{r/buildapc} (very distinctive stories).
In Figure \ref{figure:categories-distinctiveness} in  \ref{appendix-section-additional-results}, we show the full ranking of subreddit categories, which can aid researchers in determining whether to build a custom detector for their target domain or rely on a mixture of finetuning data to align with the diverse storytelling in some categories.

\section{Case Study: \textit{r/ChangeMyView}}
\label{section:cmv}

We demonstrate \datasetName's usefulness not only when measuring storytelling across communities but also when measuring storytelling across topics within a single community, by sharing a focused case study of how one subreddit uses storytelling as a rhetorical strategy.

\textit{r/ChangeMyView} is a forum dedicated to good faith debate, in which posters share opinions, commenters compete to persuade the posters to change their view, and posters assign awards for any successful persuasion.
Commenters use various rhetorical strategies, including storytelling, to persuade the poster, and this community has frequently been the subject of research about persuasion \citep{falk-lapesa-2023-storyarg}. 
Using the Winning Arguments Corpus \citep{tan2016winning}, we explore how storytelling is used to persuade.

\begin{figure}[t]
    \centering
        \centering
        \includegraphics[width=\linewidth]{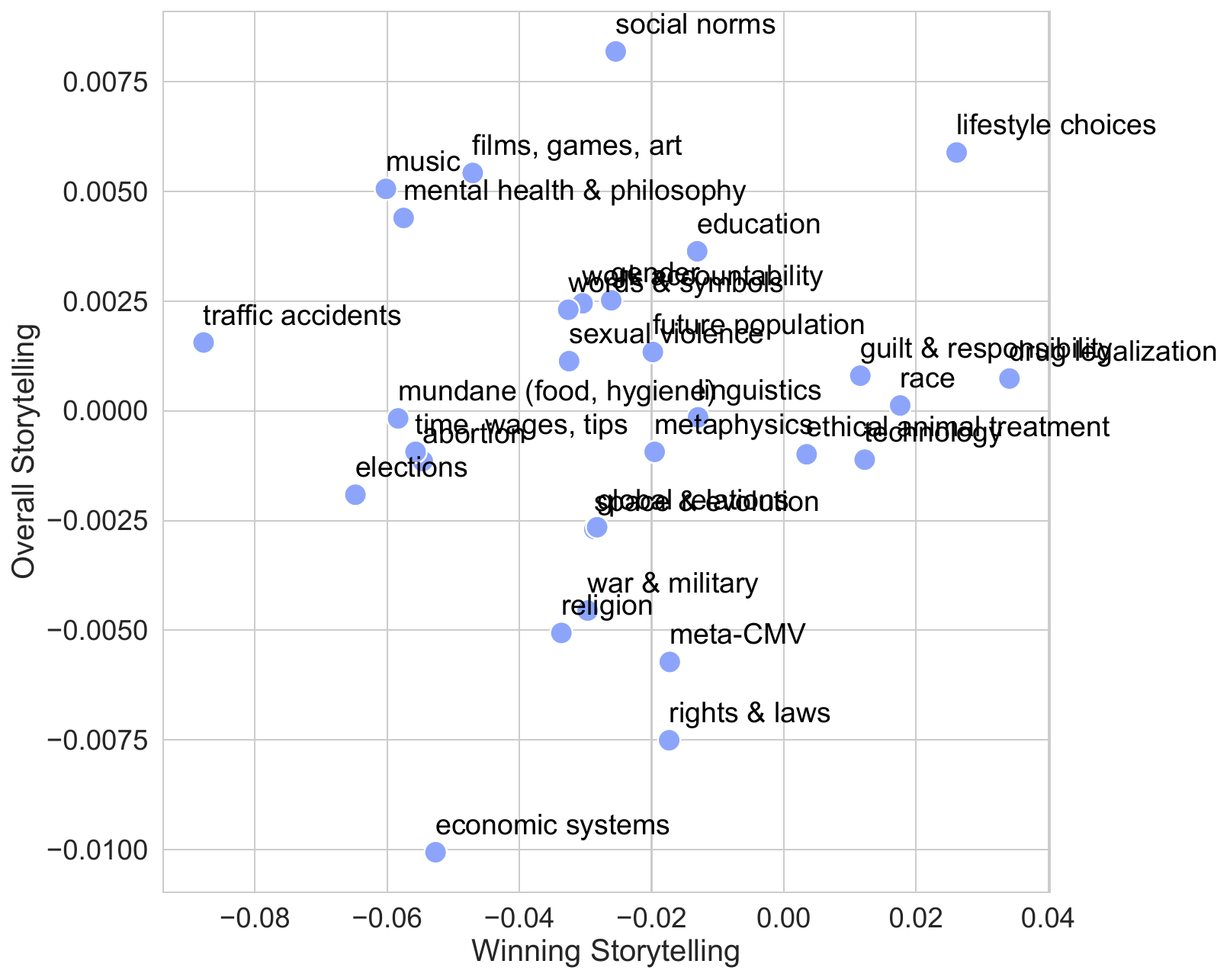}
    \caption{The debate topics plotted by the overall rate of storytelling arguments they receive (y-axis) vs. the rate of storytelling in winning comments (x-axis). 
    Some post topics, like \textit{music}, elicit many storytelling comments but very few of those comments are persuasive, while other topics like \textit{race} elicit fewer stories but responses containing stories more often tend to be persuasive. See Figure \ref{figure:cmv} in \ref{appendix-section-additional-results} for detailed views of both axes with standard deviation bounds.}
    \label{figure:cmv-scatterplot}
\end{figure}

Figure \ref{figure:cmv-scatterplot} shows the post topics that elicit comments with more or less storytelling and whose storytelling comments are more or less likely to receive a winning ``delta'' point, indicating that the poster was convinced by the comment's argument.
Topics are trained on the posts via a latent Dirichlet allocation (LDA) \citep{blei2003latent} model ($k=30$), and after examination of the highest probability words and documents, we assign each topic a descriptive name. 
The y-axis shows the difference between the mean post topic probabilities for storytelling and non-storytelling comments, while the x-axis shows the difference between the proportion of winning comments that include storytelling and the proportion of non-winning comments that include storytelling.
More details about our methods and the topic model are given in \ref{appendix-section-topic-model-cmv}.

We find that social and personal topics like \textit{lifestyle choices} and \textit{music} receive more storytelling comments, while abstract topics like \textit{economic systems} receive fewer storytelling comments.
However, topics that receive more storytelling comments are not necessarily the same topics where storytelling comments are more likely to win delta points.
Examining the outliers in Figure \ref{figure:cmv-scatterplot}, we find that for topics like \textit{drug legalization} and \textit{words \& symbols}, storytelling comments are more likely to be persuasive even though commenters are less likely to use storytelling in their arguments.
Conversely, topics like \textit{social norms} receive many storytelling comments but those comments are less likely to win. 
Our results add fine-grained distinctions to prior work that has examined ``personal and anecdotal'' arguments on \textit{r/ChangeMyView} \citep{poulsen2023large}.

These patterns likely are driven not only by the topic's subject matter but also the particular framing used in this subreddit; topics like \textit{ethical animal treatment} could be imagined to elicit to many stories, but do not in this community.

\section{Discussion}

\paragraph{Implications for narratology}
The features that are most strongly associated with storytelling support prior work's emphasis on storytelling's basis in agent-centered, event-driven forms of communication that are grounded in concrete settings. 
These features are believed to support social coordination by fostering joint attention around virtual events  \cite[known as the ``deictic theory'' of storytelling; ][]{piper2022toward}.

Nevertheless, considering narratology's historical focus on long-form literary forms like the novel, \citet{elinor_ochs_living_2009}'s famous assertion that the ``mundane conversational narratives of personal experience constitute the prototype of narrative activity rather than the flawed byproduct of more artful and planned narrative discourse'' highlights the opportunities for narrative theorists to attend to ``small stories'' \citep{georgakopoulou2007small}, such as social media stories profiled here. We hope that the \datasetName codebook, dataset, and models can serve as a cross-disciplinary bridge for narrative theorists, computational social scientists, and NLP researchers interested in large-scale analysis of narrativity across social media.

\paragraph{Implications for storytelling in online communities}
Compared to prior work that detects storytelling in online healthcare communities \citep{ganti-etal-2022-narrative,ganti-findings-2023}, our model predicts an overall \textit{lower} rate of storytelling, likely owing to the diversity of topics in our dataset.
Our classification performance is lower than the performance reported in those works, despite using similar fine-tuning techniques, signalling the challenge of identifying stories across diverse topics and communities.
While storytelling may be a sign of trust as a form of self-disclosure \citep{ma2019when}, we do not find a significant relationship between the overall storytelling rate in a specific community and the same community's  toxicity (measured via the Perspective API\footnote{\url{https://perspectiveapi.com/}}), size (number of members), or user activity (posts per member).
Finally, we provide some additional experiments about overall post and comment patterns in \ref{section:what-conversational-context}.

\section{Conclusion}

We formalize the task of story detection and story span detection by releasing \datasetName, a framework that contains a dataset, codebook, and fine-tuned models for online story detection.
Our analysis showcases how these tools can be used both across many diverse Reddit communities and to probe a specific community. 
Using our annotated story spans, we provide the first results indicating how story spans differ from non-storytelling spans cooccurring in the same texts, and we map where people tell stories.
We hope our tools and analysis spark further research into online storytelling.

\section{Limitations}

The Webis-TLDR-17 dataset provided coherent texts across a range of topics, communities, rhetorical goals, and time periods --- important qualities for our study --- but it also comes with limitations.
It only includes texts that include a ``TL;DR'' summary statement, and it only includes data through 2016.
Running our story detection system over a larger dataset would allow us to (a) study chronological patterns in storytelling and (b) study more fine-grained conversational dynamics, given the full post and comment threads contextualizing each target text.
However, our initial experiments shown in Figure \ref{figure:barplot-sd-subreddits} in the Appendix, which were sampled from the full set of posts in those subreddits, indicates that our rankings are reliable beyond the Webis-TLDR-17 dataset.

In addition, our dataset and analysis are restricted to English-language texts on Reddit.
An analysis of cultural patterns in storytelling would be important follow-up work to our study and would require an expansion across different languages.
Likewise, analyses of and comparisons across other online forums and social media platforms could help designers in understanding user behavior. 

Our annotated dataset is small, with only 500 annotated texts.
However, we emphasize the length of these texts (each text can contain up to 500 tokens) as well as the arduous nature of our annotation task, which involved multiple levels of annotation (both event and story spans), an extensive codebook with many edge cases, the need to use subjective interpretation for the annotation task, and multiple discussions to arrive at high quality consensus labels.
Despite these challenges, we achieved inter-annotator reliability scores in what is traditionally understood as the ``substantial agreement'' for both events and stories, but this required significant time and effort.

Our annotation procedure can include multiple contiguous spans of story spans for a given text. 
When a story is interrupted by non-story text, we highlight the story spans and do not highlight the interrupting non-story span. 
We observed many cases like this in the dataset, emphasizing the importance of our span annotations (rather than labeling the entire text with a single binary label).
However, we do not capture whether non-contiguous spans are part of the same story. 
Future work could augment our annotations with such story span coreference information; this would be a valuable addition to our dataset.

\section{Ethical Considerations}

Online forums like Reddit often contain toxic, explicit, and sensitive text.
For example, texts can include calls for violence, ethnic slurs, sexually explicit discussion, and private health information.
Depending on their exposure, these texts can harm both their readers (annotators, researchers) and/or their authors, if they did not intend their texts to be shared out of their original context.

While we share Reddit IDs and their corresponding annotations produced in this study,  we do not share replications of user IDs, post or comment texts, or other user information.
The post and comment IDs can be used to ``rehydrate'' the document annotations, and we release spans upon request.
The \datasetName models can be used without downloading any data.

All of the texts are used in our automatic analysis, but we attempt to remove the most potentially harmful texts from our annotators. 
After manually categorizing the subreddits (see \S\ref{subsection:data-source}), we  filter a set of  categories from the annotation task: \textit{banned}, \textit{children}, \textit{confessions}, \textit{mental health}, \textit{NSFW}, \textit{queer culture}, \textit{relationships}, \textit{drugs}, \textit{gender}, \textit{sexuality}, and \textit{toxic}.
Our goal is to (1) protect the annotators from toxic content and (2) protect the Reddit users from having their sensitive information used in an annotation task.
We also hand-select specific subreddits to filter from across the other categories (a full list is given in our public code repository).
We additionally remove 8 texts by hand from our annotated dataset after filtering; these texts were toxic, violent, and/or explicit but were posted in subreddits that we had not filtered.
We do not omit this data from our analysis but only from our annotated dataset.

All texts quoted in this article are paraphrased amalgamations of texts in our dataset; this avoids revealing information publicly that was shared in the context of a specific community, and it preserves the ability of Reddit users to edit or delete their texts.

Our study was considered exempt by the IRB at the Allen Institute for AI.

\bibliography{custom,anthology}

\clearpage
\newpage
\appendix
\onecolumn

\section{Appendix}
\label{section:appendix}

\subsection{Error Analysis}
\label{appendix-subsection-error-analysis}

For the fine-tuned RoBERTa model, we observe the following categories of errors, using open coding to categorize false positives and false negatives.

\textit{Stories misclassified as non-stories} sometimes contain cognitive verbs, such as ``plan,'' ``decide,'' or ``notice,'' which we annotated as events, using context to decide whether the verb met our criteria for specificity and sequentiality.
This category also includes stories made up of hypothetical verbs (we annotated these only when they were strongly storylike, but their occurrence is rare) and very short stories (one sentence or less).

\textit{Non-stories misclassified as stories} often contain general or repeating events or describe a state without sequence.
These texts often include pronouns, entities, and concrete language like place descriptions, making the texts appear more story-like.
These mistakes reflect some of the many edge cases that our codebook was designed to avoid but whose sparsity and ambiguity make it difficult for automatic methods to capture.

\subsection{Additional Information About Dataset}
\label{appendix-section-additional-info-dataset}

\begin{table}[H]
    \centering
    \scriptsize
    \begin{tabular}{@{}p{1.9cm}p{5.3cm}@{}}
    \toprule
    \textbf{Category} & \textbf{Example Subreddits} \\
    \midrule
    gaming & \textit{r/leagueoflegends}, \textit{r/gaming}, \textit{r/DotA2} \\[1ex]
    hobbies & \textit{r/poker}, \textit{r/photography}, \textit{r/MakeupAddiction} \\[1ex]
    tech & \textit{r/technology},  \textit{r/linux}, \textit{r/AndroidQuestions} \\[1ex]
    fandom & \textit{r/asoiaf}, \textit{r/doctorwho}, \textit{r/StarWars} \\[1ex]
    general & \textit{r/AskReddit}, \textit{r/pics}, \textit{r/Showerthoughts} \\[1ex]
    informative & \textit{r/explainlikeimfive}, \textit{r/math}, \textit{r/RealEstate} \\[1ex]
    news \& politics & \textit{r/PoliticalDiscussion}, \textit{r/changemyview}, \textit{r/Economics} \\[1ex]
    professional advice & \textit{r/legaladvice}, \textit{r/AskDocs}, \textit{r/graphic\_design} \\[1ex]
    professional sports & \textit{r/nfl}, \textit{r/hockey}, \textit{r/LiverpoolFC} \\[1ex]
    relationships & \textit{r/relationships}, \textit{dating\_advice}, \textit{r/Parenting} \\[1ex]
    fitness & \textit{r/running}, \textit{r/bodybuilding}, \textit{r/keto} \\[1ex]
    religion & \textit{r/atheism}, \textit{r/Christianity}, \textit{r/DebateReligion} \\[1ex]
    stories & \textit{r/tifu}, \textit{r/TalesFromRetail}, \textit{r/Dreams} \\[1ex]
    cities & \textit{r/toronto}, \textit{r/Seattle}, \textit{r/LosAngeles} \\[1ex]
    countries & \textit{r/canada}, \textit{r/japan} \\[1ex]
    drugs & \textit{r/Drugs}, \textit{r/LSD}, \textit{r/opiates} \\[1ex]
    mental health & \textit{r/depression}, \textit{r/ADHD}, \textit{r/socialanxiety} \\[1ex]
    queer culture & \textit{r/lgbt}, \textit{r/ainbow}, \textit{r/bisexual} \\[1ex]
    gender & \textit{r/TwoXChromosomes}, \textit{r/AskMen}, \textit{r/AskWomen} \\[1ex]
    self help & \textit{r/introvert}, \textit{r/GetMotivated}, \textit{r/INTP} \\[1ex]
    software dev & \textit{r/programming}, \textit{r/cscareerquestions}, \textit{r/gamedev} \\[1ex]
    confessions & \textit{r/offmychest}, \textit{r/DoesAnybodyElse}, \textit{r/confession} \\[1ex]
    finance & \textit{r/personalfinance}, \textit{r/Bitcoin}, \textit{r/investing} \\[1ex]
    academic & \textit{r/EngineeringStudents}, \textit{r/college}, \textit{r/GradSchool} \\[1ex]
    addiction & \textit{stopdrinking}, \textit{stopsmoking}, \textit{cripplingalcoholism} \\[1ex]
    animals & \textit{r/Pets}, \textit{r/Dogtraining}, \textit{r/cats} \\[1ex]
    \textit{healthcare} & \textit{r/BabyBumps}, \textit{r/diabetes}, \textit{r/SkincareAddiction} \\[1ex]
    other & banned subreddits, subreddits about children, NSFW and toxic subreddits, miscellaneous \\[1ex]
    \bottomrule
    \end{tabular}
    \caption{The subreddit categories and examples of member subreddits. These categories were developed via an open-coding approach followed by consolidation. Categories are shown in order of descending frequency by number of member subreddits.}
    \label{table:subreddit-categories}
\end{table}

\subsection{Stories, Toics, and Winning Arguments in \textit{r/ChangeMyView}}
\label{appendix-section-topic-model-cmv}

We trained a latent Dirichlet allocation (LDA) topic model \citep{blei2003latent} on 3,046 posts from \textit{r/ChangeMyview} distributed as part of the Winning Arguments Corpus \citep{tan2016winning}.
While newer topic modeling methods like BERTopic have become popular \citep{grootendorst2022bertopic}, LDA's performance on human coherence evaluation tests is still very strong if not stronger \citep{harrando-etal-2021-apples,hoyle-etal-2022-neural}.
We removed punctuation, normalized numbers, lower-cased the text, removed duplicate documents \citep{schofield-etal-2017-quantifying}, and did not stem or remove stop words \citep{schofield-mimno-2016-comparing,schofield-etal-2017-pulling}.
After training we examined the highest probability words and documents for each topic to qualitatively assign a label to each topic.
We show the final set of 30 topics in Table \ref{table:cmv_topics}.

\begin{figure}[H]
    \centering
        \centering
        \includegraphics[width=0.8\linewidth]{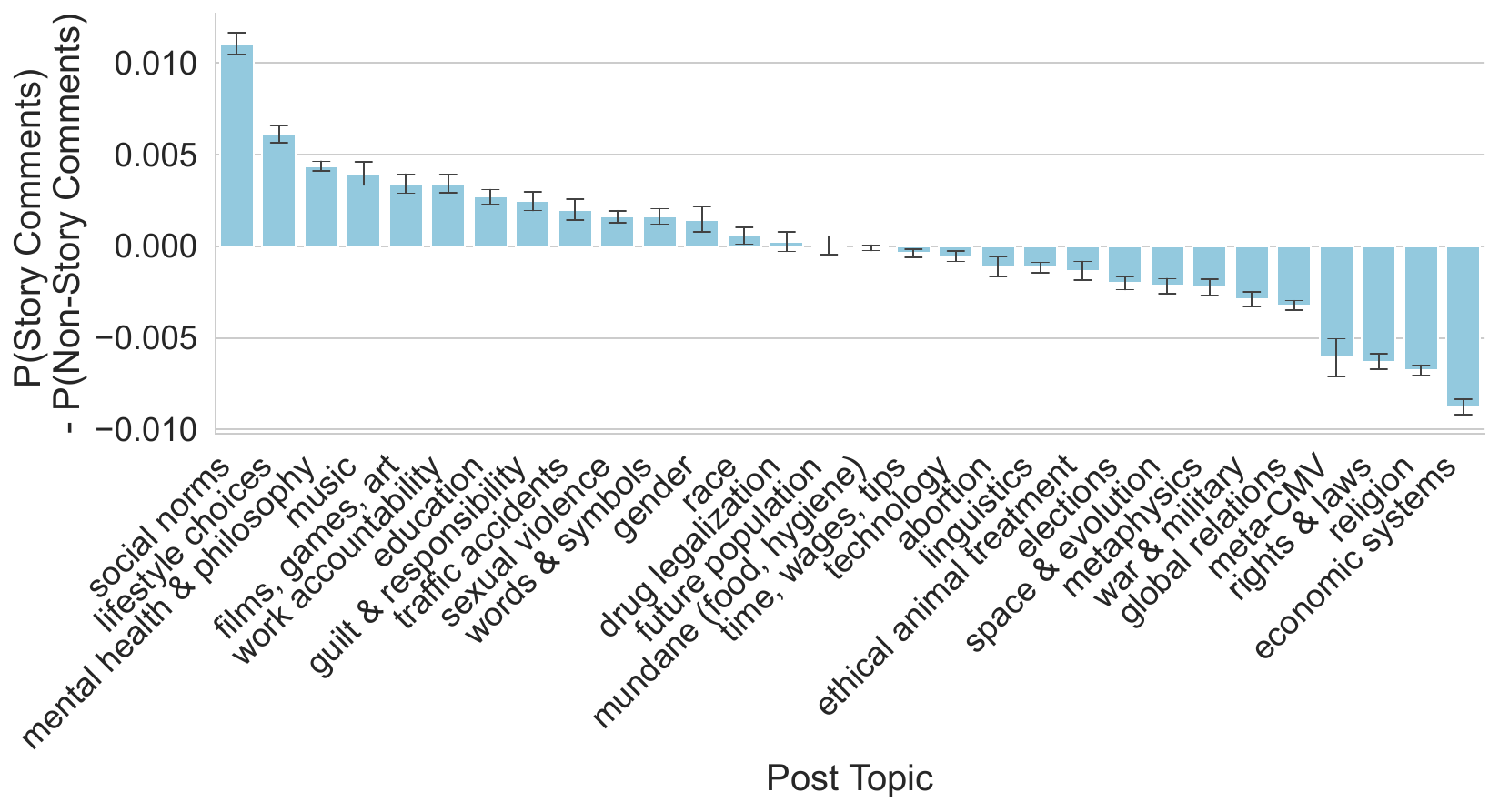}
        \includegraphics[width=0.8\linewidth]{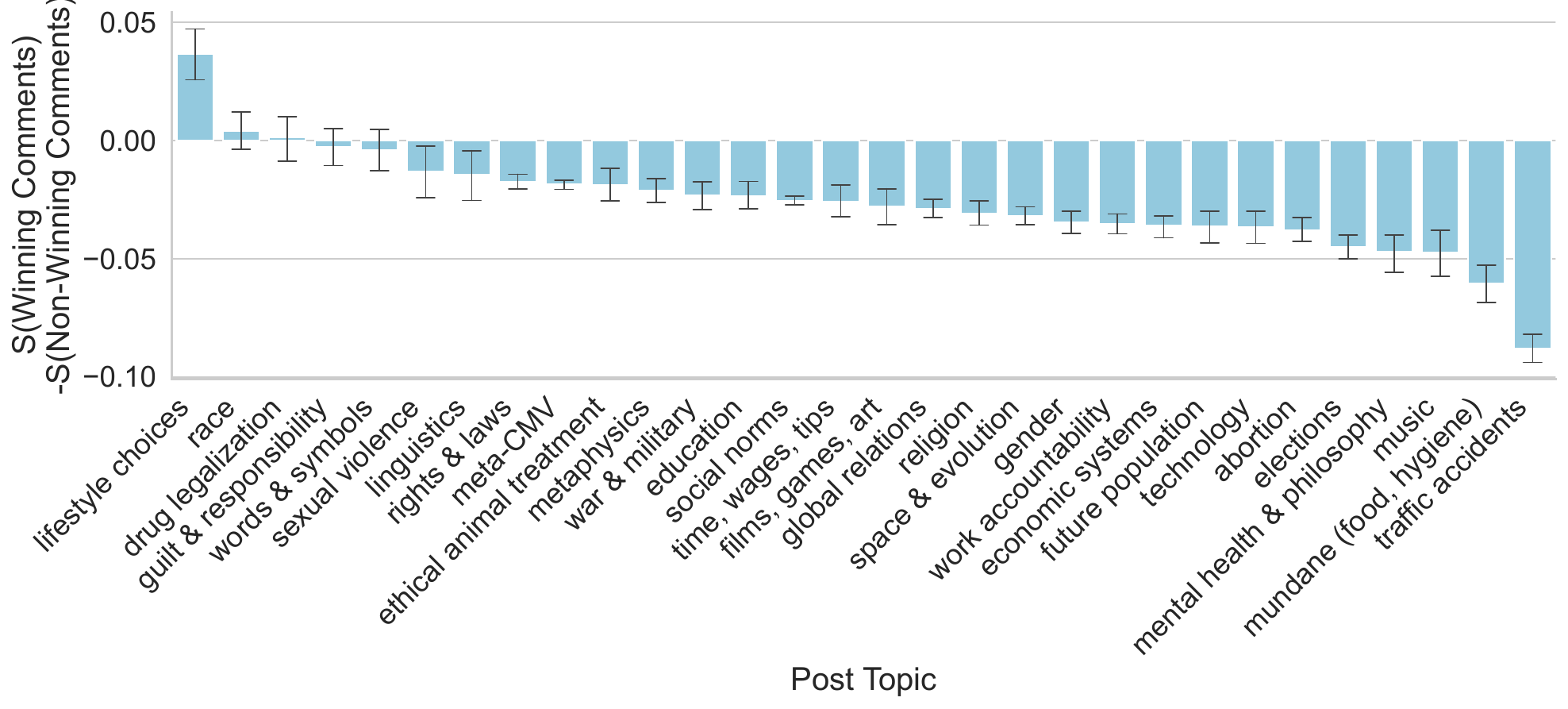}
    \caption{The first plot shows the difference in mean topic probabilities between story and non-story comments for each post topic; higher bars indicate more overall storytelling for that topic. The second plot shows the difference in proportions of winning (delta-awarded) comments containing stories and losing comments containing stories; higher bars indicate more winning comments containing storytelling.}
    \label{figure:cmv}
\end{figure}

We calculate two metrics to rank the post topics, shown by the bar plots in Figure \ref{figure:cmv}.
In each case we are ranking post topics by a measurement on comments responding to post topics.
The first metric measures \textit{overall storytelling}; we calculate the mean topic probability for all posts that storytelling comments respond to, and we subtract the the mean topic probability for all posts that non-storytelling comments respond to.
The second  metric measures \textit{winning storytelling}; for each topic, we find all the posts topic probability over 0.1, and we find the proportion of storytelling comments for that post that win a delta point and subtract the proportion of non-storytelling comments that win a delta point.

\begin{table}[H]
    \centering
    \scriptsize
    \begin{tabular}{@{}p{0.3cm}p{2.7cm}p{5cm}p{7cm}@{}}
    \toprule
    \textbf{Topic} & \textbf{Label} & \textbf{Highest Probability Tokens} & \textbf{Example Post Text} \\
    \midrule
    0 & traffic accidents & car, driving, cars, NUM, drive, traffic, stop, risk, dog, seat & Take a look at these statistics from Wikipedia: \&gt; According to the U.S. Natio... \\
1 & drug legalization & death, drugs, crime, society, health, alcohol, drug, illegal, consent, life & The official death toll in the Mexican drug war states that sixty thousand peopl... \\
2 & economic systems & money, would, pay, people, government, work, tax, income, business, job & \#\# Section I: Why is Basic Income Increasingly Popular? "Basic income" is a poli... \\
3 & war \& military & war, us, country, states, government, military, gun, united, guns, countries & Let me be as clear as possible: I’m talking only about the Army, the main branch... \\
4 & time, wages, tips & NUM, time, year, sports, number, hour, every, team, three, football & Here is an outline of my reasoning with MS paint diagrams: http://i.imgur.com/nY... \\
5 & ethical animal treatment & animals, food, meat, eat, eating, animal, humans, vegetarian, healthy, diet & Given that livestock consume more water, feed, create more greenhouse gasses, an... \\
6 & mundane (food, hygiene) & power, water, would, use, workers, gt, demand, paper, could, employers & Benn Wyatt made several good points about the virtues of calzones. For those tha... \\
7 & linguistics & culture, use, like, term, language, way, change, often, people, words & Some cases in point: - The US/English pronunciation of the name Rothschild as "R... \\
8 & education & school, college, education, students, job, schools, work, student, high, learn & Note: Although I do hold this opinion, it feels prejudiced to me so I am genuine... \\
9 & guilt \& responsibility & people, person, someone, bad, think, kill, even, self, good, lives & Batman's strict no-killing policy has lead to the deaths of hundreds of people. ... \\
10 & mental health \& philosophy & like, reddit, think, read, much, hate, seems, pretty, etc, mean & When discussing people in altered states, including those brought about through ... \\
11 & rights \& laws & right, rights, law, laws, free, gt, freedom, case, legal, argument & I believe the modern Libertarian (as defined by people like Ron Paul) is hypocri... \\
12 & future population & NUM, us, years, world, new, time, gt, better, population, means & *“Will robots inherit the earth? Yes, but they will be our children.” - Marvin M... \\
13 & work accountability & people, would, work, go, get, life, time, could, much, day & I understand that child rearing is important, but how can I have equal respect f... \\
14 & films, games, art & game, games, art, play, video, movie, movies, show, character, characters & I believe 3D is a cancer upon the film industry for the following reasons: * 3D ... \\
15 & lifestyle choices & even, many, go, end, ever, less, age, years, real, least & I have been drinking various craft beers for more than 20 years now. In the past... \\
16 & technology & use, ads, phone, buy, used, price, computer, store, internet, apple & You probably get this a lot, but I was thinking about it in the shower today. Ye... \\
17 & abortion & child, children, parents, kids, abortion, life, mother, birth, pro, family & Both the man and woman are equally responsible for an unplanned pregnancy. My re... \\
18 & music & music, great, even, time, like, sound, much, value, many, english & Why do sound technicians ruin so many indoor gigs by turning the whole volume up... \\
19 & sexual violence & rape, victim, information, issues, victims, google, internet, see, reddit, sexual & The argument from miracles is the argument that there have been miracles which a... \\
20 & social norms & like, people, get, want, really, know, see, re, think, would & This is something that's always bothered me, all the way back to when I was a ki... \\
21 & race & white, black, people, race, racism, racist, news, group, media, social & I had a heated ideological debate last night with 2 sociologists and a feminist ... \\
22 & global relations & society, many, state, world, believe, social, major, countries, science, philosophy & It appears to me that when people talk about the rise of China as a global force... \\
23 & metaphysics & one, believe, think, based, evidence, matter, also, understanding, true, know & First of all, the "you" you identify with is probably the summation of biologica... \\
24 & elections & vote, people, political, would, system, voting, party, politics, democracy, politicians & People who have no interest in politics, and are not interested in learning abou... \\
25 & gender & women, men, sex, gender, male, gay, woman, man, female, sexual & Gender, as defined by Wikipedia, is "the range of characteristics pertaining to,... \\
26 & words \& symbols & word, use, definition, gt, using, used, logic, said, first, considered & EDIT: Please refrain from making anymore comments about controlled demolitions, ... \\
27 & meta-CMV & view, edit, people, think, would, change, changed, believe, point, post & Lately (especially since the invasion from /r/adviceanimals), there have been a ... \\
28 & space \& evolution & would, one, could, life, likely, make, police, environment, might, human & First off, let me clarify, I am not defending the actions of Ferguson Police Off... \\
29 & religion & god, religion, human, religious, believe, moral, good, would, beliefs, belief & The typical Christian resolution to the problem of evil is to state that it is h... \\
    \bottomrule
    \end{tabular}
    \caption{The 30 topics derived from a topic model trained the posts in \textit{r/ChangeMyView}. We show the 10 words with highest probability as well as the prefix of an example post text for each topic.}
    \label{table:cmv_topics}
\end{table}

\subsection{In what conversational contexts do people tell stories?}
\label{section:what-conversational-context}

Prior theoretical work has emphasized the importance of turn-taking and interaction among actors as a key determinant of narrative behavior \cite{georgakopoulou2007small}. 
According to this paradigm, stories
depend on the social interactions that elicit and modulate their telling \citep{herman2009basic}.
Similarly, prior work on trust in online communities \citep{galegher1998legitimacy} suggests that self-disclosures, such as personal stories, occur more frequently in communities where trust is higher \citep{yang2019channel,ma2017self,joinson2010privacy}, an important metric of community health.

As an initial foray, we find that \textit{posts} are more likely to contain stories than \textit{comments} across most communities, with a mean ratio of storytelling rates in posts versus comments of 2.28. 
Some subreddits that ranked very low for overall storytelling nevertheless have relatively high rates of storytelling in posts; e.g., \textit{r/askscience} has an overall low storytelling rate (0.03) but ranks highest for storytelling in posts versus comments (0.56 in posts, 0.09 in comments). 
Question-asking subreddits can be found at both ends of the ranking (e.g., \textit{r/NoStupidQuestions} has a high ratio of storytelling in posts versus comments while \textit{r/AskReddit} has a low ratio), and via a Pearson correlation test, we do not find a significant correlation between rates of question-asking (i.e., the rate of question mark characters) and storytelling in posts ($p>0.05$).
More results are in Table \ref{table:post-comment-ratios}.

\begin{table}[H]
    \centering
    \scriptsize
    \begin{tabular}{@{}p{3cm}p{1cm}p{1cm}p{1cm}@{}}
    \toprule
    \textbf{Subreddit} & \textbf{Ratio} & \textbf{$P(s|p)$} & \textbf{$P(s|c)$} \\
    \midrule
    \multicolumn{4}{c}{\textit{Subreddits with \textbf{highest} post:comment storytelling ratio}} \\
    \midrule
    \textit{r/askscience} & 6.35 & 0.56 & 0.09 \\[1ex]
    \textit{r/philosophy} & 3.83 & 0.36 & 0.10 \\[1ex]
    \textit{r/legaladvice} & 3.58 & 0.97 & 0.27 \\[1ex]
    \textit{r/NoStupidQuestions} & 3.47 & 0.68 & 0.19 \\[1ex]
    \textit{r/summonerschool} & 3.40 & 0.62 & 0.18 \\[1ex]
    \textit{r/LeagueofLegendsMeta} & 3.38 & 0.41 & 0.12 \\[1ex]
    \textit{r/Bitcoin} & 3.08 & 0.53 & 0.17 \\[1ex]
    \textit{r/applehelp} & 3.07 & 0.77 & 0.25 \\[1ex]
    \textit{r/techsupport} & 3.02 & 0.82 & 0.27 \\[1ex]
    \textit{r/poker} & 2.88 & 0.77 & 0.27 \\[1ex]
    \midrule
    \multicolumn{4}{c}{\textit{Subreddits with \textbf{lowest} post:comment storytelling ratio}} \\
    \midrule
    \textit{r/nfl} & 1.21 & 0.50 & 0.41 \\[1ex]
    \textit{r/LifeProTips} & 1.21 & 0.62 & 0.52 \\[1ex]
    \textit{r/weddingplanning} & 1.20 & 0.92 & 0.77 \\[1ex]
    \textit{r/TalesFromRetail} & 1.19 & 0.97 & 0.82 \\[1ex]
    \textit{r/DoesAnybodyElse} & 1.16 & 0.77 & 0.66 \\[1ex]
    \textit{r/travel} & 1.16 & 0.76 & 0.66 \\[1ex]
    \textit{r/harrypotter} & 1.16 & 0.81 & 0.70 \\[1ex]
    \textit{r/AskReddit} & 1.10 & 0.88 & 0.80 \\[1ex]
    \textit{r/SquaredCircle} & 1.10 & 0.70 & 0.64 \\[1ex]
    \textit{r/Random\_Acts\_Of\_Amazon} & 1.01 & 0.90 & 0.89 \\[1ex]
    \bottomrule
    \end{tabular}
    \caption{Ranking of the subreddits by storytelling in posts versus comments. Also shown are the probabilities of storytelling $s$ given either a post $p$ or comment $c$.}
    \label{table:post-comment-ratios}
\end{table}

\begin{figure}[H]
    \centering
        \centering
        \includegraphics[width=0.5\linewidth]{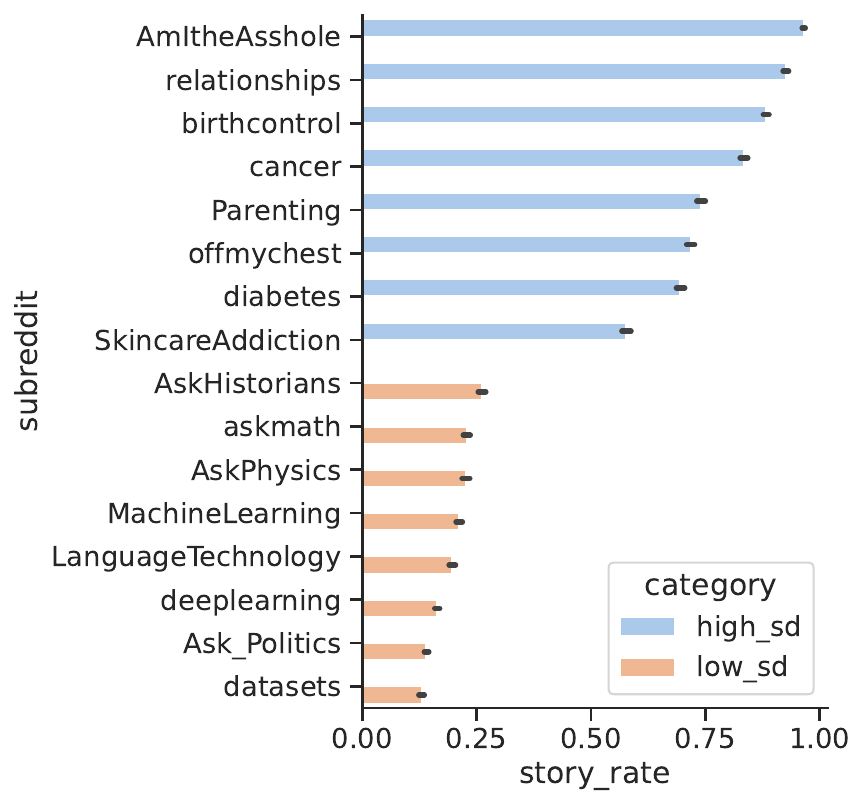}
    \caption{Subreddits with expected \textit{high} or \textit{low} rates of self-disclosure and their predicted storytelling rates, using the fine-tuned RoBERTa model and our \textit{consensus} annotations. As expected, subreddits with higher rates of self-disclosure also have higher rates of storytelling. In this plot, we are showing results for the a set of 500 posts randomly sampled from each subreddit rather than limiting our analysis to \textit{coherent} posts from the Webis-TLDR dataset, providing validation that our other ranksings are consistent.}
    \label{figure:barplot-sd-subreddits}
\end{figure}

Finally, over a targeted set of eight subreddits expected to have high rates of self-disclosure (\textit{relationships}, \textit{healthcare}) and eight subreddits expected to have low rates of self-disclosure (\textit{technical questions}, \textit{machine learning}), we confirm that the high self-disclosure subreddits also contain more storytelling, according to predictions generated by our fine-tuned RoBERTa model.
Results are shown in Figure \ref{figure:barplot-sd-subreddits}.

\subsection{Additional Results}
\label{appendix-section-additional-results}

\begin{figure}[H]
    \centering
        \centering
        \includegraphics[width=0.5\linewidth]{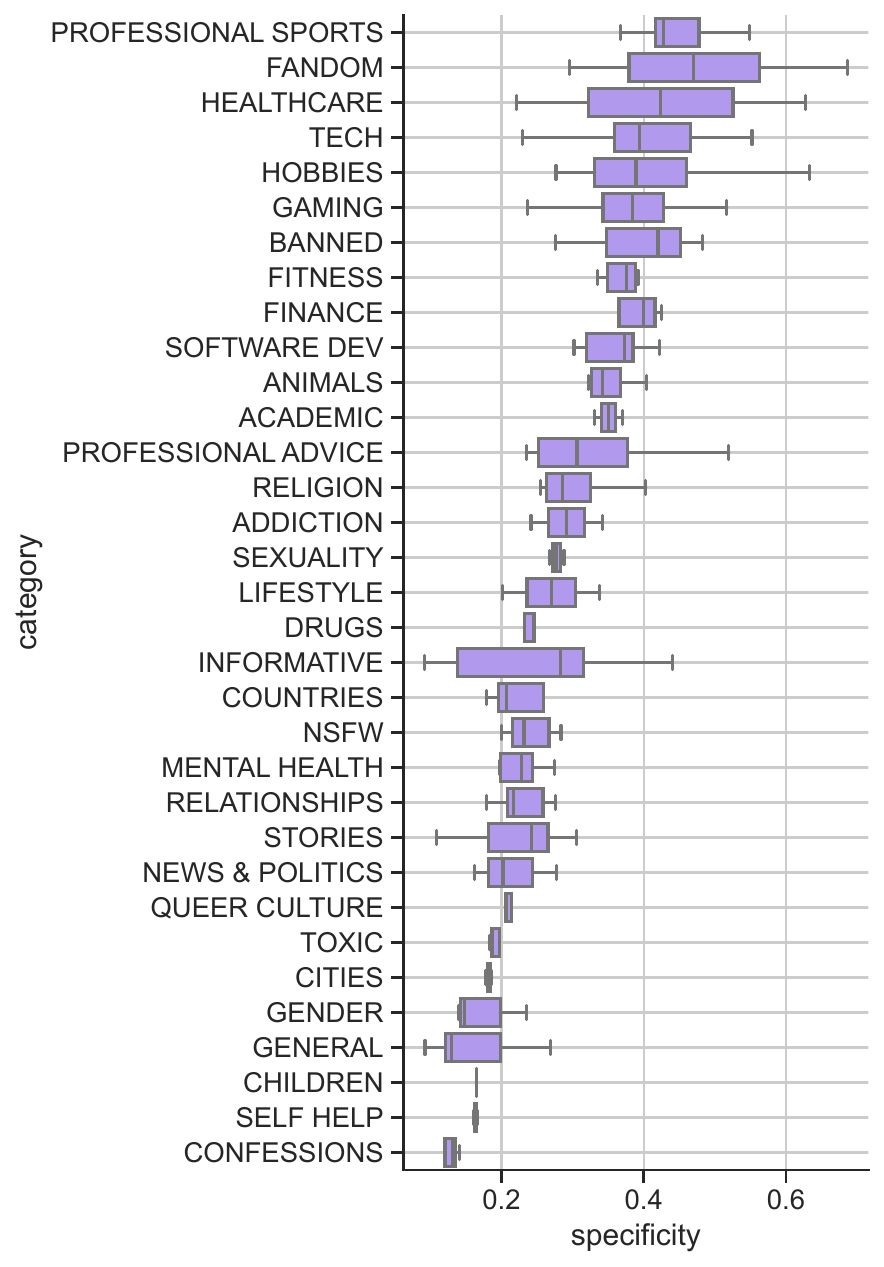}
    \caption{The subreddit categories ranked by their distinctiveness scores.}
    \label{figure:categories-distinctiveness}
\end{figure}

\begin{figure}[H]
    \centering
        \centering
        \includegraphics[width=0.28\linewidth]{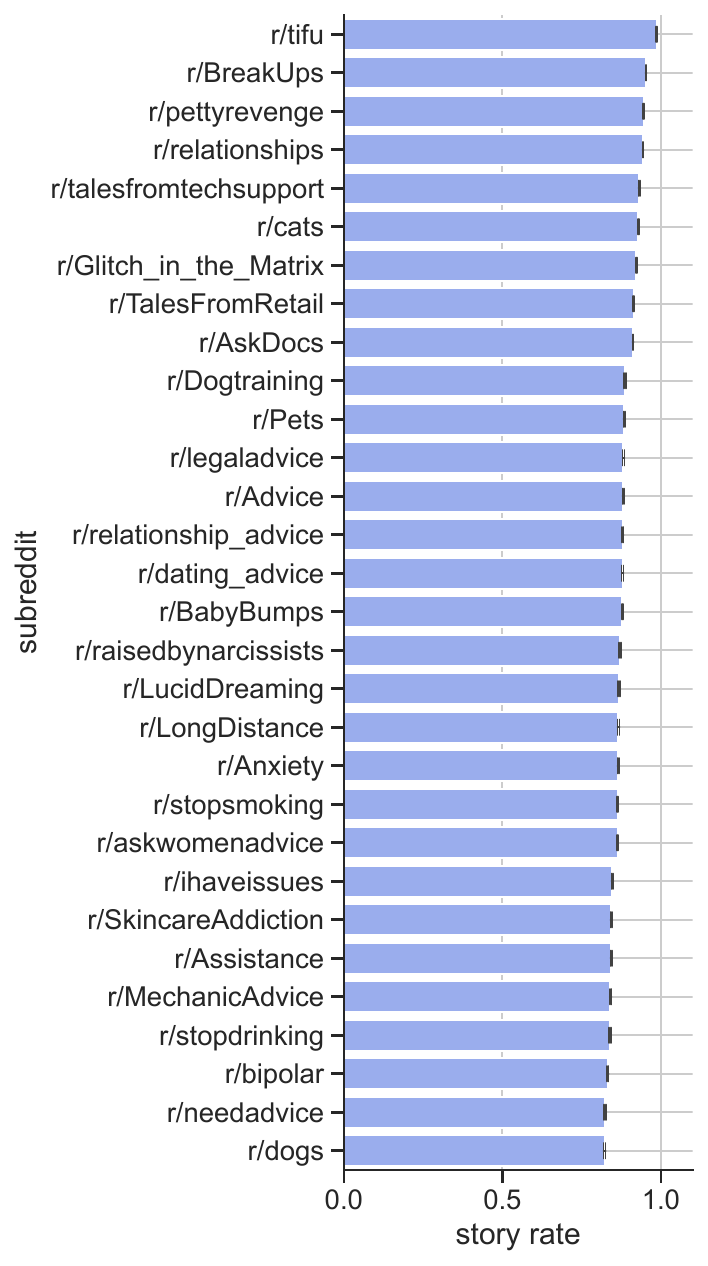}
        \includegraphics[width=0.29\linewidth]{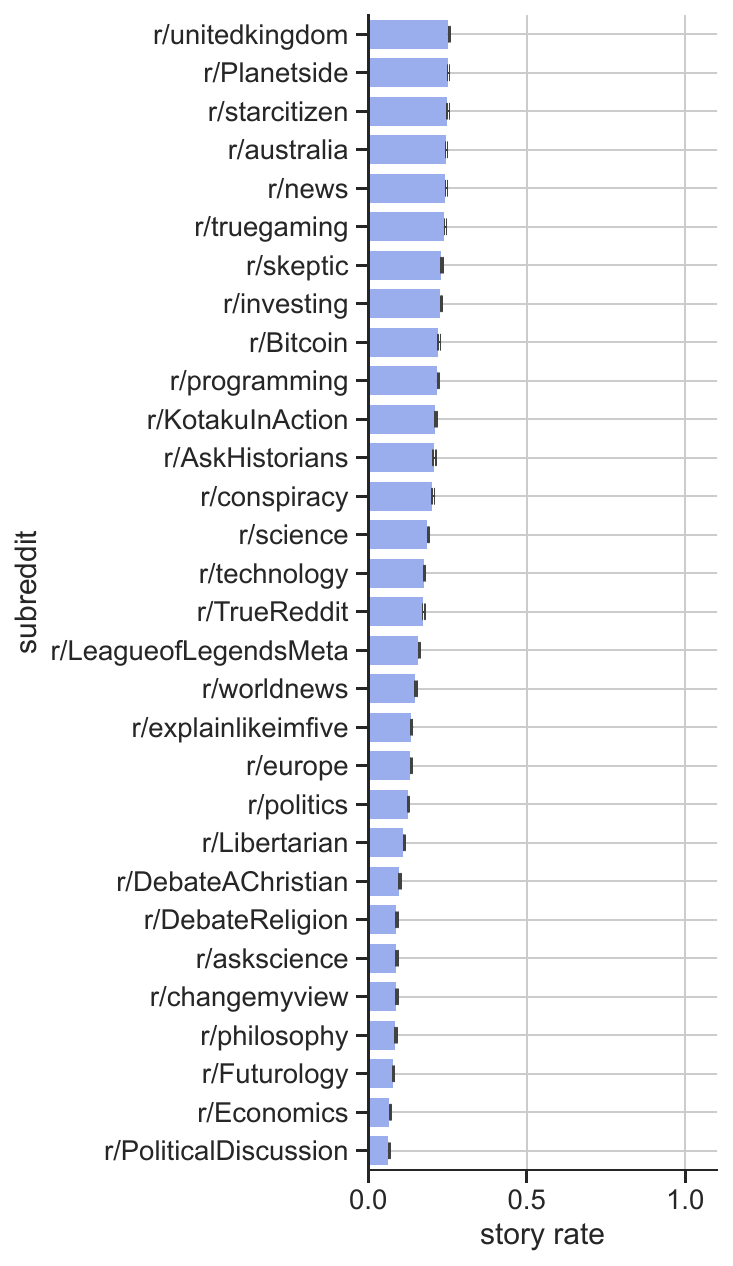}
        \includegraphics[width=0.35\linewidth]{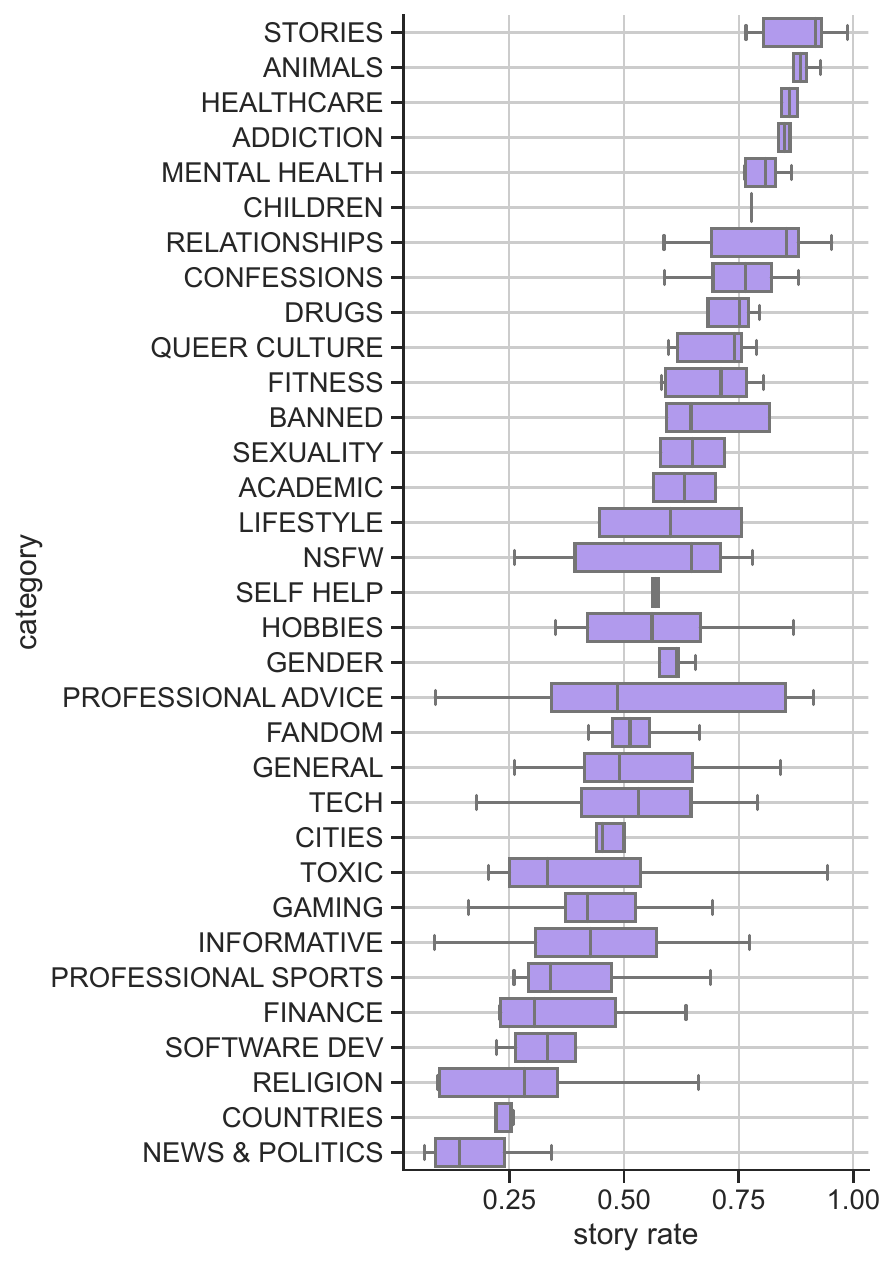}
    \caption{Subreddits (left) and categories of subreddits (right) ranked by their rate of texts (posts and comments) containing stories, as predicted by \datasetName. Results represent 20 bootstrapped samples of the texts for each subreddit.}
    \label{figure:boxplot-categories}
\end{figure}

\begin{table}[H]
    \centering
    \scriptsize
    \begin{tabular}{@{}p{1cm}|p{2.5cm}p{2.5cm}@{}}
    \toprule
    & \textbf{Less Storytelling} & \textbf{More Storytelling}  \\
    \midrule
    \textbf{Generic}
    & \textit{r/politics} \newline
      \textit{r/explainlikeimfive} \newline
      \textit{r/PoliticalDiscussion} \newline
      \textit{r/Futurology}
    & \textit{r/tifu} \newline
      \textit{r/pettyrevenge} \newline
      \textit{r/Glitch\_in\_the\_Matrix} \newline
      \textit{r/Advice}
    \\[1ex]
    \textbf{Distinctive}
    & \textit{r/asoiaf} \newline
      \textit{r/summonerschool} \newline
      \textit{r/Naruto} \newline
      \textit{r/fantasyfootball}
    & \textit{r/SkincareAddiction} \newline
      \textit{r/LucidDreaming }\newline
      \textit{r/techsupport} \newline
      \textit{r/MechanicAdvice}
    \\[1ex]
    \bottomrule
    \end{tabular}
    \caption{Subreddits with the most or least storytelling and most or least distinctive vocabulary.}
    \label{table:matrix-specificity-storytelling}
\end{table}

\subsection{Prior Definitions of Stories}
\label{appendix-subsection-prior-definitions}

The following story definitions are drawn from prior work in NLP.

\smallskip
\noindent
\textit{``A narrative is a discourse presenting a coherent sequence of events which are
causally related and purposely related, concern specific characters and times, and
overall displays a level of organization beyond the commonsense coherence of the
events themselves, such as that provided by a climax or other plot structure.''} -- \citet{Eisenberg2021Narrative}

\smallskip
\noindent
\textit{``A narrative is a discourse presenting a coherent sequence of events which are causally related and purposely related, concern specific characters and times, and overall displays a level of organization beyond the commonsense coherence of the events themselves. In sum, a story is a series of events effected by animate actors… at least two key elements to stories, namely, the plot (fabula) and the characters (dramatis personae) who move the plot forward (Abbott, 2008).''} -- \citet{eisenberg-finlayson-2017-simpler}

\smallskip
\noindent
\textit{``It is generally agreed in narratology (Forster, 1962; Mani, 2012; Pentland, 1999; Bal, 2009) that a narrative presents a sequence of events arranged in their time order (the plot) and involving specific characters (the characters).''} -- \citet{yao-huang-2018-temporal}

\smallskip
\noindent
\textit{``A narrative is the recounting of a sequence of events that have a continuant subject and constitute a whole (Prince, 2003).''} -- \citet{castricato-etal-2021-fabula}

\smallskip
\noindent
\textit{``A sequence of related events, leading to a resolution or projected resolution.''} -- \citet{ceran2012semantic}

\smallskip
\noindent
\textit{``Situatedness: narrativity depends on the social context in which it occurs. Event sequencing: narrativity depends on temporally ordered events. World making: narrativity depends on the fact of disequilibrium such that we can observe a change in the world.''} -- \citet{piper2021detecting}

\smallskip
\noindent
\textit{``Operationalizing  \citet{smith2001discourse}'s  characteristics,  our  codebook had four inclusion criteria: the presence of a plot, characters, the author as a character, and a clear beginning, middle,and end. Additionally, we included three exclusion criteria. Posts were marked as non-narrative that were: purely informational, entirely providing resources, or entirely composed of a question posed to the subreddit community.''} -- \citet{doyle2024stories}

\smallskip
\noindent
\textit{``A STORY follows a plot-like structure (e.g. has an introduction, middle section or conclusion) or contains a sequence of events.''} -- \citet{falk-lapesa-2023-storyarg}

\begin{table}[H]
    \centering
    \scriptsize
    \begin{tabular}{@{}p{3cm}p{4cm}@{}}
    \toprule
    \textbf{Features Used in Definition} & \textbf{Prior Work} \\
    \midrule
    sequences of events arranged temporally & \citet{piper2021detecting} \newline \citet{yao-huang-2018-temporal} \newline \citet{castricato-etal-2021-fabula} \newline \citet{doyle2024stories} \newline \citet{falk-lapesa-2023-storyarg} \\[1ex]
    causally related events leading to resolutions & \citet{eisenberg-finlayson-2017-simpler} \newline \citet{ceran2012semantic} \newline \citet{alzahrani2016story} \\[1ex]
    entities or characters & \citet{eisenberg-finlayson-2017-simpler} \newline \citet{piper2021detecting} \newline \citet{yao-huang-2018-temporal} \newline \citet{alzahrani2016story} \newline \citet{doyle2024stories} \\[1ex]
    rhetorical purpose & \citet{eisenberg-finlayson-2017-simpler} \newline \citet{roos2021narratives} \newline \citet{castricato-etal-2021-fabula}  \\[1ex]
    world building or setting & \citet{piper2021detecting}  \\[1ex]
    \bottomrule
    \end{tabular}
    \caption{Features used in story \textbf{definitions} from prior work.}
    \label{table:story-features-prior-work}
\end{table}

\begin{table}[H]
    \centering
    \scriptsize
    \begin{tabular}{@{}p{3cm}p{4cm}@{}}
    \toprule
    \textbf{Features Used for Prediction} & \textbf{Prior Work} \\
    \midrule
    n-gram & \citet{dos2017portuguese} \newline \citet{piper2021detecting} \newline \citet{gordon2009identifying} \newline \citet{ceran2012semantic} \\[1ex]
    part of speech & \citet{yao-huang-2018-temporal} \newline \citet{piper2021detecting} \newline \citet{ceran2012semantic} \\[1ex] 
    coreference chain length & \citet{eisenberg-finlayson-2017-simpler} \newline \citet{yao-huang-2018-temporal} \\[1ex]
    LIWC & \citet{dos2017portuguese} \newline \citet{yao-huang-2018-temporal} \\[1ex]
    readability & \citet{dos2017portuguese} \\[1ex]
    verb classes & \citet{eisenberg-finlayson-2017-simpler} \\[1ex]
    syntactic production rules & \citet{yao-huang-2018-temporal} \\[1ex]
    verb sequence perplexity & \citet{yao-huang-2018-temporal} \\[1ex]
    dependency tags & \citet{piper2021detecting} \\[1ex]
    tense & \citet{piper2021detecting} \\[1ex]
    mood & \citet{piper2021detecting} \\[1ex]
    voice & \citet{piper2021detecting} \\[1ex]
    amount of dialog & \citet{piper2021detecting} \\[1ex]
    named entities & \citet{ceran2012semantic} \\[1ex]
    stative verbs & \citet{ceran2012semantic} \\[1ex]
    semantic triples & \citet{ceran2012semantic} \\[1ex]
    \bottomrule
    \end{tabular}
    \caption{Examples of features used to \textbf{predict} storytelling in prior work.}
    \label{table:story-features-prior-work}
\end{table}

\subsection{Full Codebook}
\label{appendix-section-full-codebook}

This codebook drew from \citet{sims-etal-2019-literary} and \citet{piper2021detecting} with significant modifactions for our online setting, story detection task, and span-based annotation.

\smallskip
\noindent
\textbf{Does this text contain a story?}
\textit{Use the guidelines below to support your decisions, but ultimately, follow your best judgment as there are many edge cases.}

\smallskip
\noindent
\textbf{A story describes a sequence of events involving one or more people.}
\begin{itemize}
    \item Stories can be fictional or real, exciting or mundane.
    \item Focus only on the current text. Don’t worry about whether there might be a story before or after this text. References to stories aren’t stories.
    \item Stories describe the experiences of one or more specific people.
    \begin{itemize}
        \item “People” can include animals, aliens, etc.
        \item “People” can include groups as long as these are specific groups of people that exist at a specific time and place.
        \item “People” includes the first person narrator.
    \end{itemize}
    \item Stories must include multiple, specific events. 
    \begin{itemize}
        \item These events should be sequential: one event happens, then another event happens. It’s ok if the events are narrated out of order, but there should still be a clear sequence.
        \item These events should be connected: they might be about the same people, they might be causally connected, they might describe an overall change or transformation in the state of the world, they might describe a single experience.
        \item Jumbles of events that are unordered and/or unconnected (like lists of examples) are not stories.
    \end{itemize}
    \item What are events?
    \begin{itemize}
        \item Events are “a singular occurrence at a particular place and time.”
        \item General, repeating, isolated, or hypothetical situations, states, and actions are usually not events, unless they appear together in a strongly story-like sequence. 
        \item Most stories are told in the past tense. Present and future tense can also be used, but the bar is higher and the narrated events need to be strongly story-like.
        \item Most events are positively asserted as occurring, but depending on the context, negative verbs can also be events when occurring at a specific time and place.
        \begin{itemize}
            \item For example: ``I tried to leave the room, but the door wouldn't open.''
            \item For example: ``He asked her to make cookies for his birthday today, but she didn't make him cookies because she doesn't have an oven at home.'' (``didn't make'' is an event but ``doesn't have'' is not an event)
        \end{itemize}
        \item Events are usually verbs but can also be nouns and adjectives.
        \item When are states events? See \citet{sims-etal-2019-literary}.
    \end{itemize}
    \item When highlighting any spans:
    \begin{itemize}
        \item Do not highlight spans in the post title.
    \end{itemize}
    \item When highlighting the story spans:
    \begin{itemize}
        \item Include all the text you think is part of the story. This should include not just events but also text that sets the stage, summarizes the story, ends with a lesson learned, etc.
        \item Text that usually shouldn’t be included in the story span:
        \begin{itemize}
            \item introductory text about the subreddit, why they’re posting, etc.
            \item questions about the story
            \item explanations, discussion, hypotheses external to the story
        \end{itemize}
        \item Ask yourself: Is this text necessary if I were writing a summary of the story?
    \end{itemize}
    \item When highlighting the event spans:
    \begin{itemize}
        \item Highlight only one word per event. For example, highlight only “am” in the phrase "am walking slowly".
        \item As a rule, never highlight infinitives
        \item The ``ing'' form of a verb should usually never be highlighted if it is acting as a noun (e.g. ``Demanding forgiveness is unfair''). If it is acting as an adjective (e.g. ``His demanding look'') it may or may not be highlighted, depending on context. If it is acting as a present participle in an event’s verb phrase (e.g. ``They are demanding a response''), then highlight ``are''.
    \end{itemize}
\end{itemize}

\subsection{GPT Prompts}
\label{appendix-section-gpt-prompts}

\subsubsection{Few-Shot Document Classification}

We modify the below prompt based on the setting. For zero-shot, we exclude the ``Examples'' section. We test two few-shot settings: $k=2$ and $k=4$ (where $k$ is the number of examples). In each case, we sample from the training dataset until we have a total of $k$ examples, evenly split between positive (i.e. story) and negative (i.e. not story) instances from the training dataset. Because we use k-fold cross validation, we sample examples from the fold-specific training dataset. We then interleave the negative and positive samples in the ``Examples'' section.

We also try excluding the Guidelines section. We find that including the Guidelines section improves performance in the zero-shot setting. In contrast, the best performing few-shot setting for each model used $k=4$ examples and excluded the Guidelines section.

\begin{quote}
\setlength{\parindent}{0cm}\ttfamily{
Guidelines: \newline \newline
Use the guidelines below to support your decisions, but ultimately, follow your best judgment as there are many edge cases.}

\setlength{\parindent}{0cm}\ttfamily{
A story describes a sequence of events involving one or more people.}

\setlength{\parindent}{0cm}\ttfamily{
Stories can be fictional or real, exciting or mundane. Stories describe the experiences of one or more specific people. “People” can include animals, aliens, etc. “People” can include groups as long as these are specific groups of people that exist at a specific time and place. “People” includes the first person narrator.}

\setlength{\parindent}{0cm}\ttfamily{
Stories must include multiple, specific events. These events should be sequential: one event happens, then another event happens. It’s ok if the events are narrated out of order, but there should still be a clear sequence. These events should be connected: they might be about the same people, they might be causally connected, they might describe an overall change or transformation in the state of the world, they might describe a single experience. Jumbles of events that are unordered and/or unconnected (like lists of examples) are not stories.}

\setlength{\parindent}{0cm}\ttfamily{
Events are “a singular occurrence at a particular place and time.” General, repeating, isolated, or hypothetical situations, states, and actions are usually not events. Most stories are told in the past tense. Present and future tense can also be used, but the bar is higher and the narrated events need to be strongly story-like. Most events are positively asserted as occurring, but depending on the context, negative verbs can also be events when occurring at a specific time and place. Events are usually verbs but can also be nouns and adjectives.}

\setlength{\parindent}{0cm}\ttfamily{
Examples: \newline\newline
Text: <TEXT>\newline
Answer: <YES/NO>\newline
}

\setlength{\parindent}{0cm}\ttfamily{
Task:\newline\newline
A story describes a sequence of events involving one or more people. Does the following text contain a story? Answer yes or no, and then explain your reasoning.\newline\newline
Text: \newline
Answer: \newline
}
\end{quote}

\subsubsection{Chain-of-Thought Document Classification}

We use the following prompt template to present a simplified breakdown of the classification task into two subtasks: identifying qualifying characters and identifying qualifying events.

\begin{quote}
\setlength{\parindent}{0cm}\ttfamily{
Your task is to decide whether a text contains a story. You should follow the guidelines below and think step by step.
}

\setlength{\parindent}{0cm}
\ttfamily{Use the guidelines below to support your decisions, but ultimately, follow your best judgment as there are many edge cases.}

\setlength{\parindent}{0cm}
\ttfamily{A story describes a sequence of events involving one or more people.} 

\setlength{\parindent}{0cm}
\ttfamily{Stories can be fictional or real, exciting or mundane. Stories describe the experiences of one or more specific people. “People” can include animals, aliens, etc. “People” can include groups as long as these are specific groups of people that exist at a specific time and place. “People” includes the first person narrator.}

\setlength{\parindent}{0cm}
\ttfamily{Stories must include multiple, specific events. These events should be sequential: one event happens, then another event happens. It’s ok if the events are narrated out of order, but there should still be a clear sequence. These events should be connected: they might be about the same people, they might be causally connected, they might describe an overall change or transformation in the state of the world, they might describe a single experience. Jumbles of events that are unordered and/or unconnected (like lists of examples) are not stories.}

\setlength{\parindent}{0cm}
\ttfamily{Events are “a singular occurrence at a particular place and time.” General, repeating, isolated, or hypothetical situations, states, and actions are usually not events. Most stories are told in the past tense. Present and future tense can also be used, but the bar is higher and the narrated events need to be strongly story-like. Most events are positively asserted as occurring, but depending on the context, negative verbs can also be events when occurring at a specific time and place. Events are usually verbs but can also be nouns and adjectives.}

\setlength{\parindent}{0cm}
\ttfamily{Using the definitions for `people' and `events' given in the guidelines above, answer the following questions}

\setlength{\parindent}{0cm}
\ttfamily{Question 1: Does the text contain `people'? If so, list them.}

\setlength{\parindent}{0cm}
\ttfamily{Question 2: Does the text contain a sequence of causally connected `events'. If so, list them.}

\setlength{\parindent}{0cm}
\ttfamily{Finally, based on the guidelines and your answers to Question 1 and Question 2, decide whether the text contains a story. Respond `Yes' or `No'.}

\setlength{\parindent}{0cm}
\ttfamily{Two examples are provided below:}

\setlength{\parindent}{0cm}
\ttfamily{Example 1:
`Yesterday, I went to the store to buy a jug of milk so that I could make pancakes. When I arrived, the cashier told me that the store lost power the previous night, so all the milk spoiled. So much for pancakes!'}

\setlength{\parindent}{0cm}\ttfamily{
 Question 1 Answers: [`I', `cashier'].\newline 
 Question 2 Answers: [`went', `arrived', `told', `lost', `spoiled'].\newline 
 Story Decision: Yes}

\setlength{\parindent}{0cm}
\ttfamily{
Example 2:\newline
`Not a good FDIC tip. Coverage levels start at \$250k and can be increased by certain multipliers--so that's not really a concern unless you've got several million in the bank.'}

\setlength{\parindent}{0cm}
\ttfamily{
Question 1 Answers: []. \newline
Question 2 Answers: []. \newline
Story Decision: No \newline\newline
Your turn. \newline
Text: <TEXT>}
\end{quote}

\subsubsection{Few-Shot Story Boundary Detection}
Formulating a story boundary detection task for GPT-4 is difficult, due the proclivity for responses to include unrequested mutations to the original source string, which can lead to token misalignment. We mitigate this issue through careful prompt engineering, including manually-constructed few-shot examples that demonstrate desired behavior on different formats observed in our data, and setting the temperature to 0 in OpenAI chat completion requests. The results support the token-level story span detection results (see Table \ref{table:classification-performance}). We include the prompt below. Finally, we note that in consideration of the poor performance of this method relative to the Fine-tuned RoBERTa model, in general we do not recommend this approach for for token-level discourse boundary detection tasks.

\begin{quote}
\setlength{\parindent}{0cm}\ttfamily{
A story describes a sequence of events involving one or more people. Stories can be fictional or real, exciting or mundane. Stories describe the experiences of one or more specific people. “People” can include animals, aliens, etc. “People” can include groups as long as these are specific groups of people that exist at a specific time and place. “People” includes the first person narrator. Stories must include multiple, specific events. These events should be sequential: one event happens, then another event happens. It’s ok if the events are narrated out of order, but there should still be a clear sequence. These events should be connected: they might be about the same people, they might be causally connected, they might describe an overall change or transformation in the state of the world, they might describe a single experience. Jumbles of events that are unordered and/or unconnected (like lists of examples) are not stories. Events are “a singular occurrence at a particular place and time.” General, repeating, isolated, or hypothetical situations, states, and actions are usually not events. Most stories are told in the past tense. Present and future tense can also be used, but the bar is higher and the narrated events need to be strongly story-like. Most events are positively asserted as occurring, but depending on the context, negative verbs can also be events when occurring at a specific time and place. Events are usually verbs but can also be nouns and adjectives.}

\setlength{\parindent}{0cm}\ttfamily{
Stories may or may not span the entire text. Below are some guidelines for determining what belongs in a story span.
1. Story spans include all of the text you think is part of the story. This should include not just events but also text that sets the stage, summarizes the story, ends with a lesson learned, etc.
2.  Parts of texts that usually do not belong to a story span include introductory text about the subreddit, why they’re posting, etc; questions about the story; explanations, discussion, hypotheses external to the story
3. Text that would be necessary to summarize the story usually belongs in the story span.
}

\setlength{\parindent}{0cm}\ttfamily{
The text below may or may not contain one or more stories. If it does contain one or more stories, those stories may or may not span the entire text. Your task is to annotate any story spans in the text. To mark the start of a story span, insert <<S>>. To mark the end of a story, insert <<E>>. If you insert an <<S>>, you must eventually insert an <<E>>. The first marker you insert must always be <<S>> and you may never insert an <<E>> unless the last marker you inserted was an <<S>>.
}

\setlength{\parindent}{0cm}\ttfamily{
Aside from optionally inserting <<S>> and <<E>> markers, you must return the entire original text exactly as it was presented to you. Even if the input text starts with a title in brackets or HTML tags (e.g. ['Title' or '<b>Title:</b>') or has line breaks, HTML tags, grammatical errors, or random sequences of characters, you must not make unauthorized changes and output the entire original text, including any title. This is the most important rule you must follow no matter what. Think carefully before your respond to make sure you are following my instructions.}

\setlength{\parindent}{0cm}\ttfamily{
Example 1:
[Title: Forgot my lunch at home]
}

\setlength{\parindent}{0cm}\ttfamily{
Hi everyone, this is my first post here, so please be nice. Anyway, yesterday I forgot my lunch at home. I am a picky eater, so I decided to buy a snack from a vending machine rather than eat the cafeteria food at the office. One of my coworkers made a comment, which I thought was rude. What do you all think?
}

\setlength{\parindent}{0cm}\ttfamily{
Example 1 Answer:
[Title: Forgot my lunch at home]
}

\setlength{\parindent}{0cm}\ttfamily{
Hi everyone, this is my first post here, so please be nice. <<S>>Anyway, yesterday I forgot my lunch at home. I am a picky eater, so I decided to buy a snack from a vending machine rather than eat the cafeteria food at the office. One of my coworkers made a comment, which I thought was rude.<<E>> What do you all think?
}

\setlength{\parindent}{0cm}\ttfamily{
Example 2:
<b>Comment:</b>Does anyone have major political disagreements with a close family member? Thanksgiving is coming up and I'm curious how people in this situation are planning to handle awkward conversations.
What do you all think?
}

\setlength{\parindent}{0cm}\ttfamily{
Example 2 Answer:
<b>Comment:</b>Does anyone have major political disagreements with a close family member? Thanksgiving is coming up and I'm curious how people in this situation are planning to handle awkward conversations.
}

\setlength{\parindent}{0cm}\ttfamily{
Example 3:
<b>Title:</b> I [24F] am about to fail my Bio class... again. Help. (<br><br><b>Post:</b> Idk what to do AGGHHHH!!!!?\$\$@\#. Help please. I started studying religiously two weeks ago, and I already covered 4 chapters, so maybe I'm on the right track? I talked to the professor, but he didn't offer much help. One TA sent me a list of additional resources, but it's not realistic for me to read all of those in time for the exam. 
}

\setlength{\parindent}{0cm}\ttfamily{
Example 3 Answer:
<b>Title:</b> I [24F] am about to fail my Bio class... again. Help. (<br><br><b>Post:</b>  Idk what to do AGGHHHH!!!!?\$\$@\#. Help please. <<S>>I started studying religiously two weeks ago, and I already covered 4 chapters, so maybe I'm on the right track? I talked to the professor, but he didn't offer much help. One TA sent me a list of additional resources, but it's not realistic for me to read all of those in time for the exam.<<E>> 
}

\setlength{\parindent}{0cm}\ttfamily{
Example 4:
I went to the mall yesterday and walked into Sephora for the first time in years. I knew it was expensive, but it's out of control. I ended up trying samples for a few minutes, and then caved and bought a lip liner. There went \$20. Oh well
}

\setlength{\parindent}{0cm}\ttfamily{
Example 4 Answer:
<<S>>I went to the mall yesterday and walked into Sephora for the first time in years. I knew it was expensive, but it's out of control. I ended up trying samples for a few minutes, and then caved and bought a lip liner. There went \$20. Oh well<<E>> 
}

\setlength{\parindent}{0cm}\ttfamily{
Example 5:
<b>Comment:</b>They are terrible to their customers! I truly don't understand. I went their last month and they didn't pay any attention to me until I hunted down the hostess to ask for a table. Then the food took an hour to arrive. And it wasn't even good fwiw.
}

\setlength{\parindent}{0cm}\ttfamily{
Example 5 Answer:
<b>Comment:</b><<S>>They are terrible to their customers! I truly don't understand. I went their last month and they didn't pay any attention to me until I hunted down the hostess to ask for a table. Then the food took an hour to arrive. And it wasn't even good fwiw.<<E>>
}

\setlength{\parindent}{0cm}\ttfamily{
Text:
<TEXT>
}
\end{quote}

\subsection{Additional Context on Story Feature Analysis}
\label{appendix-section-story-feature-analysis}

\subsubsection{Story Features}

\paragraph{Entities}
Entity and pronoun rates have been used in prior work to both define and detect storytelling \citep{eisenberg-finlayson-2017-simpler, piper2022toward}. 
To capture entities, we compute the proportion of several pronoun groups in the texts, including first-person singular, first-person plural, second person, and third-person singular.\footnote{first-person singular: `i', `me', `my', `myself', `mine'; first-person plural: `we', `us', `our', `ourselves', `ours'; second-person: `you', `your', `yours', `yourself', `yourselves'; third-person singular: `he', `she', `his', `her', `him', `hers', `himself', `herself'}
Additionally, we consider the entity mention rate, defined as the proportion of third-person singular pronouns plus the number of times the spaCy EntityRecognizer detects a \textit{PERSON} entity in the text.

\paragraph{Events}
Prior work has emphasized eventfulness as a key predictor of storytelling behavior \cite{huhn2009event, gius2022towards}. We consider event rates in stories based on two event detection methods. First, we use the union of the event labels from the two expert annotators. Second, following \citet{sap-pnas-2022}, we use a BERT \textit{realis} event tagger trained on a dataset of realis events in literary texts \citep{sims-etal-2019-literary}.\footnote{We adapted the BERT tagger in \url{https://github.com/maartensap/ACL2019-literary-events}. After training, the model achieved F-1 scores of 0.776 and 0.717 on the validation and test sets, respectively.}
These metrics allow us to compare our event definition to past event definitions and how these interact with our story labels. 

\paragraph{Verb Tense}
Previous research has suggested that temporal distance is a key function in establishing the state of joint attentionality \cite{tomasello2010origins} among narrator and audience members \cite{piper2022toward}. To evaluate the importance of verb tense, we sort verbs into two groups: past-tense and not past-tense. Our heuristic for detecting verb tense is based on a partition of the six Penn Treebank \citep{marcus-etal-1993-building} verb subtypes employed by the spaCy part-of-speech tagger. We assign \textit{VBD} and \textit{VBN} to the past tense group and all other verb tags to complementary group.

\paragraph{Concreteness}
Concreteness has been found to be a strong indicator of narrativity in books \citep{piper2022toward}. 
Our concreteness rating for texts is based on the lexicon from \citet{Brysbaert_Warriner_Kuperman_2013}. 
We take the weighted proportion of terms in the text that appear in the lexicon. 

\smallskip
\noindent
\textbf{Text type} is the text's status as post or comment, and \textbf{length} is the number of tokens in the text and the mean number of tokens in the sentences.

\subsubsection{Story Feature Comparison Across Datasets}
In \S6.1, we conduct $t$-tests comparing features between texts labeled as containing stories vs. not containing stories in the \datasetName dataset where the story group is composed solely by the story spans, as opposed to the entire text labeled as containing a story. Here, we share complementary test where full texts are used (rather than just story spans) for the story group. We perform the same tests on another narrative detection dataset, which which contains mostly literary texts \citep{piper2022toward}. 

\begin{table*}[t]
    \centering
    \scriptsize
    \begin{tabular}    {@{}p{2,9cm}|p{1.5cm}p{1.0cm}p{1.2cm}|p{1.5cm}p{1.0cm}p{1.2cm}@{}}
    \toprule
    \textbf{Measure} & \textbf{Effect Size ($d$)} & \textbf{Direction} & \textbf{$p$-value} & \textbf{Effect Size ($d$)} & \textbf{Direction} & \textbf{$p$-value}\\
    \midrule
    & \multicolumn{3}{c}{\textit{\datasetName Dataset}}  |& \multicolumn{3}{c}{\textit{Piper-Bagga Dataset}} \\
    \midrule
    expert-annotated events
    & 1.899***
    & \colorbox{story-color}{story}
    & $p<0.001$
    & n/a
    & n/a
    & n/a
    \\[1ex]
    realis events
    & 1.429***
    & \colorbox{story-color}{story}
    & $p<0.001$
    & 1.687***
    & \colorbox{story-color}{story}
    & $p<0.001$
    \\[1ex]
    past tense
    & 1.408***
    & \colorbox{story-color}{story}
    & $p<0.001$
    & 1.207***
    & \colorbox{story-color}{story}
    & $p<0.001$
    \\[1ex]
    1st-person singular pronouns
    & 1.009***
    & \colorbox{story-color}{story}
    & $p<0.001$
    & 0.84***
    & \colorbox{story-color}{story}
    & $p<0.001$
    \\[1ex]
    concreteness
    & 0.439***
    & \colorbox{story-color}{story}
    & $p<0.001$
    & 1.526***
    & \colorbox{story-color}{story}
    & $p<0.001$
    \\[1ex]
    3rd-person singular pronouns
    & 0.397***
    & \colorbox{story-color}{story}
    & $p<0.001$
    & 1.281***
    & \colorbox{story-color}{story}
    & $p<0.001$
    \\[1ex]
    entity mentions
    & 0.285**
    & \colorbox{story-color}{story}
    & 0.006
    & 1.085***
    & \colorbox{story-color}{story}
    & $p<0.001$
    \\[1ex]
    non-past tense
    & 0.947***
    & \colorbox{non-story-color}{non-story}
    & $p<0.001$
    & --
    & --
    & 0.639
    \\[1ex]
    is comment (vs. post)
    & 0.612***
    & \colorbox{non-story-color}{non-story}
    & $p<0.001$
    & n/a
    & n/a
    & n/a
    \\[1ex]
    2nd-person pronouns
    & 0.444***
    & \colorbox{non-story-color}{non-story}
    & $p<0.001$
    & 0.285*
    & \colorbox{story-color}{story}
    & 0.017
    \\[1ex]
    sentence length
    & 0.259*
    & \colorbox{non-story-color}{non-story}
    & 0.012
    & 0.828***
    & \colorbox{non-story-color}{non-story}
    & $p<0.001$
    \\[1ex]
    \textit{1st-person plural pronouns}
    & --
    & --
    & 0.106
    & --
    & --
    & 0.154
    \\[1ex]
    \textit{text length}
    & --
    & --
    & 0.106
    & 0.825***
    & \colorbox{non-story-color}{non-story}
    & $p<0.001$
    \\[1ex]
    \bottomrule
    \end{tabular}
    \caption{Results of $t$-tests comparing features between texts labeled as containing stories vs. not containing stories in the \datasetName dataset and the PiperBagga dataset, which contains mostly literary texts \citep{piper2022toward}. Because the PiperBagga dataset does not include story span annotations, we consider the entire text labeled as containing a story for the story group for the \datasetName tests, for comparison purposes. We control for multiple comparisons using the Holm method (***: $p<0.001$; **: $p<0.01$; *: $p<0.05$).}
    \label{table:story-features-2}
\end{table*}

Our findings confirm many of those found using a hand-annotated story dataset ($N=394$) from \citet{piper2022toward}, which included mostly literary and non-social media texts.
However, some differences emerged; e.g., the rate of non-past-tense is significantly lower in our stories, but there is no significant relationship in the PiperBagga dataset.

\end{document}